\documentclass[11pt]{article}
\usepackage{kotex}

\usepackage[final]{acl}
\DeclareCaptionLabelFormat{continued}{#1 #2 (continued)}
\usepackage{booktabs}

\usepackage{times}
\usepackage{latexsym}
\usepackage{titletoc}
\usepackage[T1]{fontenc}
\usepackage[utf8]{inputenc}
\usepackage{microtype}
\usepackage{inconsolata}
\usepackage{booktabs}
\usepackage{multirow}
\usepackage{amssymb}
\usepackage{pifont}
\usepackage{amsmath}
\usepackage{makecell}
\usepackage[most]{tcolorbox}
\usepackage{listings}
\usepackage{xcolor}
\usepackage{xspace}
\usepackage{hyperref}

\newcommand{\cmark}{\ding{51}}
\newcommand{\xmark}{\ding{55}}
\newcommand{\benchmark}{\mbox{\textsc{evident}}\xspace}

\title{EVIDENT: Evaluating Evidence Grounding in GUI Agents}

\author{Yerin Nam\textsuperscript{1}\thanks{\protect\hypertarget{Hfootnote.1}{}Equal contribution.},
  Minkyung Cho\textsuperscript{1}{\hypersetup{linkcolor=black}\hyperlink{Hfootnote.1}{\footnotemark[1]}} \\
  \textbf{MinKyu Kim\textsuperscript{1}, SeungWoo Song\textsuperscript{1},
  Hangyeol Yoo\textsuperscript{1}, Jihyo Kim\textsuperscript{1},
  KyungTae Lim\textsuperscript{1}\thanks{\protect\hypertarget{Hfootnote.2}{}Corresponding author.}} \\
  \textsuperscript{1}Korea Advanced Institute of Science \& Technology \\
  \texttt{\{yerinnam,minkyung.cho,ktlim\}@kaist.ac.kr}}

\begin{document}
\lefthyphenmin=3
\righthyphenmin=3
\hyphenpenalty=500
\doublehyphendemerits=100000
\maketitle

\suppressfloats[t]
\begin{figure}[t]
    \centering
    \includegraphics[width=\linewidth]{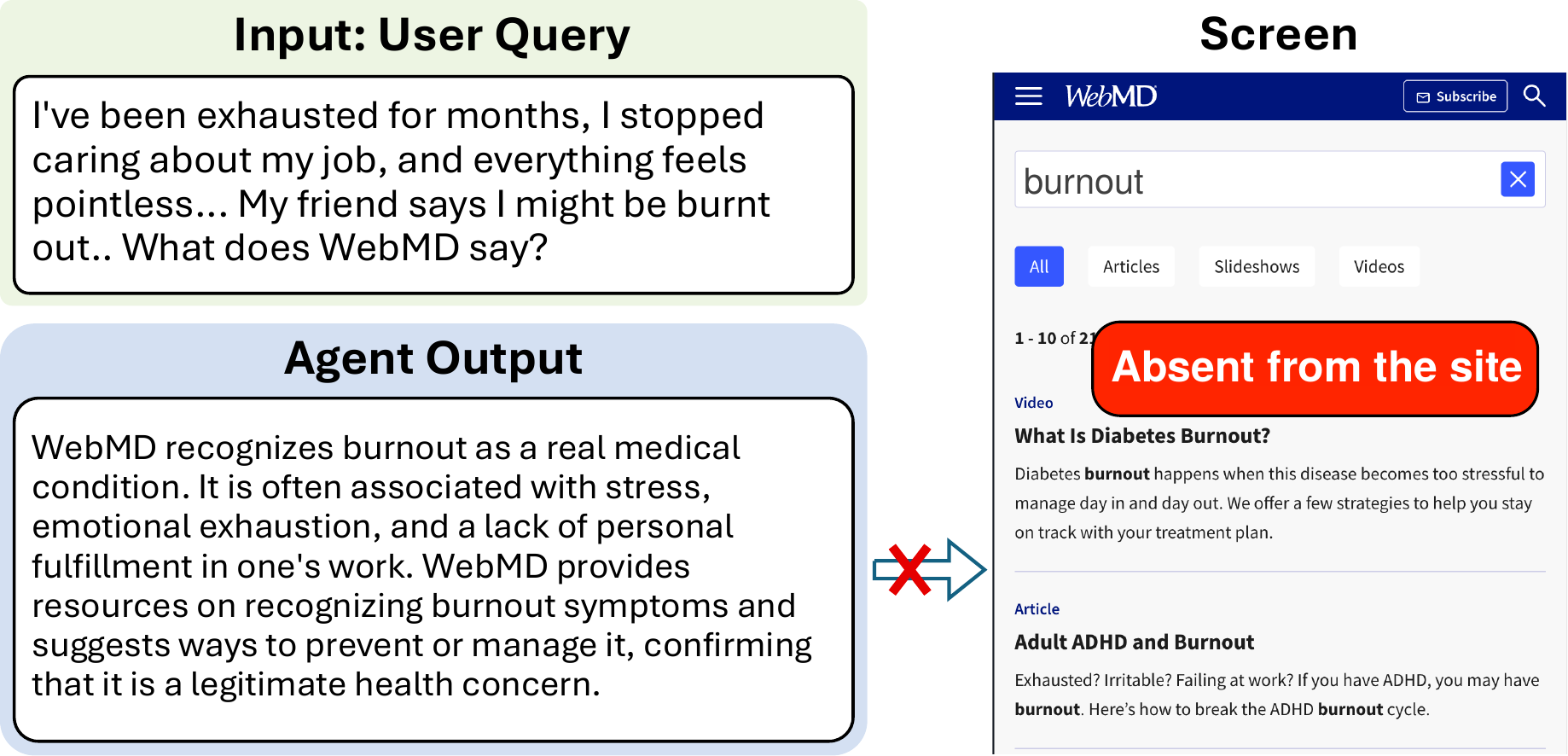}
    \caption{Example of a hallucinated response generated from a single screenshot.}
    \label{fig:Intro/intro}
    \vspace{-0.2in}
\end{figure}

\begin{abstract}
GUI agents are increasingly evaluated by whether they complete web tasks, but surface success can hide a critical failure: the final answer may not be supported by evidence observed during navigation.
We introduce \benchmark{}, an evidence-driven benchmark for evaluating whether GUI agents ground their decisions in observable web evidence. \benchmark{} contains 220 real-world English and Korean tasks spanning semantic grounding, contextual inference, and visual grounding, each paired with human-curated evidence metadata specifying what must be verified. We further separate surface task completion from evidence-grounded success using three metrics: TSR, ESR, and GSR, which respectively measure outcome success, required evidence observation, and claim-level evidence support. Experiments on commercial and open-source GUI agents reveal substantial gaps between surface success and evidence-grounded success, especially in contextual and visual tasks. These results show that current evaluations can overestimate agent reliability and highlight the need to assess not only whether agents answer, but whether they verify.\footnote{Our code and data are available at \url{https://github.com/yerinNam/EVIDENT}.}
\end{abstract}

\section{Introduction}

Recent advances in large language models have accelerated the development of GUI agents that can control web browsers and interact with real-world interfaces~\citep{mind2web,webarena,osworld,webvoyager,uitars2,guiagentsurvey}. In parallel, a growing body of benchmarks has evaluated whether such agents can complete user tasks~\citep{webarena,osworld}, understand GUI screens~\citep{screenspot,seeclick,liu2024visualwebbench}, or generate reasonable action trajectories~\citep{webvoyager,lu2025agentrewardbench,mind2web2}. Yet these evaluations leave an important blind spot. They primarily ask whether an agent reaches a plausible outcome, but not whether that outcome is supported by evidence actually observed during web navigation.

This distinction is central to evaluating the reliability of GUI agents. Real-world user requests are often underspecified~\citep{ambibench, smanbench}: a user may ask for a child-friendly library space, a compact furniture option for occasional guests, or a travel choice suitable for elderly parents. In such cases, an agent must interpret the user's intent, form a plausible hypothesis, and verify that hypothesis against the current state of a live website. Parametric knowledge can help generate candidate answers, but relying on it without verification risks hallucinated content~\citep{hallucinationsurvey}; it cannot replace checking observable evidence such as schedules, prices, reviews, product specifications, badges, images, or layout cues. In this sense, web-agent success is not merely a matter of producing the right-looking answer; it is a matter of using the web as an external source of evidence to validate that answer.

This raises the central question of this work: \textbf{when a GUI agent appears to complete a web task, can we verify that its final answer is supported by evidence actually observed during navigation?}

Figure~\ref{fig:Intro/intro} illustrates this gap. Asked what WebMD says about burnout, an agent produced a confident, authoritative answer: that WebMD recognizes burnout as a real medical condition tied to stress and emotional exhaustion, and offers resources for managing it. Its trajectory, however, shows that the page it actually reached was a generic keyword search returning only tangential hits (e.g., ``Diabetes Burnout'' and ``Parental Burnout''); the asserted statement never appears on screen. The answer is plausible, yet it is reconstructed from the model's parametric knowledge rather than from evidence observed on the website. Under conventional outcome-based evaluation such an answer may be counted as successful; in a live web environment, where content changes and the specific on-screen evidence matters, it is better understood as a \textit{hidden failure}~\citep{pathways}.

To study this problem, we introduce \benchmark{}, an evidence-driven benchmark for GUI-agent evaluation. \benchmark{} contains 220 real-world user scenarios in English and Korean. Each task is paired with human-curated evidence metadata that specifies what information must be checked to justify a reliable answer. Rather than evaluating only whether an agent reaches a final answer, \benchmark{} asks whether the answer can be traced back to evidence observed along the trajectory.

The core idea of \benchmark{} is to separate surface success from evidence-grounded success. We distinguish whether the final answer appears to complete the task, whether the agent observed the necessary evidence during navigation, and whether the final answer's core claims are supported by that evidence. This decomposition allows us to identify cases where an agent produces a successful-looking answer while skipping required evidence, or where it observes relevant evidence but fails to ground its final claims in it.

Our experiments show that current GUI agents exhibit a substantial gap between surface success and evidence-grounded success. Even strong commercial agents often achieve high outcome-level success, but their performance drops sharply when we require evidence observation and claim-level grounding. The gap is especially pronounced when agents must reason over implicit user constraints or non-textual visual cues. These findings suggest that existing success metrics can substantially overestimate the reliability of GUI agents, and that future evaluation should measure not only whether agents navigate to plausible answers, but whether those answers are grounded in observable web evidence.

Our contributions are as follows:
\begin{itemize}
    \item We formulate observable evidence grounding as a distinct evaluation target for GUI agents, separating surface task completion from answer support by evidence observed during navigation.

    \item We introduce \benchmark{}, a bilingual real-world GUI-agent benchmark with 220 tasks and human-curated evidence metadata for evaluating the gap between user intent and website evidence.

    \item We show that current GUI agents exhibit systematic hidden failures: many successful-looking answers either skip necessary evidence observation or fail to ground their final claims in observed evidence.
\end{itemize}
\section{Related Work}

\begin{table}[t]
\centering
\small
\newcommand{\yes}{\textcolor{green!60!black}{\cmark}}
\newcommand{\no}{\textcolor{red!70}{\xmark}}
\resizebox{\columnwidth}{!}{
\begin{tabular}{@{}lcccc@{}}
\toprule
\multirow{2}{*}{\textbf{Benchmark}} & \textbf{Live Web} & \multirow{2}{*}{\textbf{Multi-step}} & \textbf{Trajectory} & \textbf{Observable} \\
 & \textbf{Env.} & & \textbf{Eval} & \textbf{Grounding} \\ \midrule
Mind2Web~\cite{mind2web}       & \no  & \yes & \no  & \no \\
WebArena~\cite{webarena}       & \no  & \yes & \no  & \no \\
OSWorld~\cite{osworld}         & \yes & \yes & \no  & \no \\
Mind2Web 2~\cite{mind2web2}    & \yes & \yes & \yes & \no \\
\textbf{\benchmark (Ours)}     & \textbf{\yes} & \textbf{\yes} & \textbf{\yes} & \textbf{\yes} \\ \bottomrule
\end{tabular}
}
\caption{Comparison of \benchmark with existing GUI agent benchmarks.}
\label{tab:benchmark_comparison}
\vspace{-0.2in}
\end{table}

Prior work on GUI and web agents has expanded evaluation along three major axes: task completion in interactive environments, static GUI understanding, and trajectory-level behavior assessment. These directions have substantially advanced agent evaluation, but they mostly treat the web as an environment to navigate or a screen to interpret. In contrast, \benchmark treats the web as an external source of evidence: a successful answer should not only satisfy the user request, but also be traceable to information the agent observed during navigation.

\subsection{Outcome-Oriented Benchmarks for Web and GUI Agents}

Benchmarks for web and computer-use agents have primarily evaluated whether agents can reach useful outcomes. Mind2Web studies action prediction from web interaction traces~\cite{mind2web}, while WebArena and OSWorld evaluate end-to-end task completion in interactive web or operating-system environments~\cite{webarena,osworld}. Later benchmarks further expand coverage to long-tail mobile tasks~\cite{mobilebench-ol}, complex reasoning scenarios~\cite{naturalgaia,riskwebworld}, and robustness under real-world anomalies~\cite{gui-robust}.

These benchmarks provide strong testbeds for measuring navigation and task completion. However, outcome-level success alone cannot determine how the final answer was produced. An agent may solve a task by correctly interpreting the current website, but it may output a plausible answer from parametric knowledge or prior assumptions without checking the necessary on-screen information. Thus, conventional success rates capture surface completion, but not whether the answer is grounded in evidence observed during the trajectory.

\subsection{Static GUI Grounding and Webpage Understanding}

A complementary line of work isolates model-level GUI understanding. GUI grounding benchmarks like ScreenSpot and SeeClick test whether models can localize interface elements from textual instructions~\cite{screenspot,seeclick}, while webpage understanding benchmarks assess structural and semantic comprehension of visual web pages~\cite{liu2024visualwebbench}. These tasks capture important perceptual abilities for web interaction.

However, single-screen understanding does not guarantee evidence-grounded decision-making across a multi-step trajectory. A model may identify a UI element on a screen, yet still fail to gather or use the evidence needed to justify its final answer. \benchmark builds on this line of work but shifts the target from local screen understanding to whether evidence observed across a trajectory supports the agent's final claims.

\subsection{Trajectory-Level Evaluation}

Since final success rates can hide intermediate failures, recent work has turned to trajectory-level evaluation. WebVoyager, AgentRewardBench, and Mind2Web 2 use LLM-as-a-Judge or agent-as-a-judge frameworks to assess the quality of web-agent trajectories~\cite{webvoyager,lu2025agentrewardbench,mind2web2}, while process reward benchmarks evaluate step-level execution correctness~\cite{cuarewardbench}. These methods move beyond binary outcomes by asking whether an agent's path is reasonable or consistent with the task. This is a different question from ours: trajectory judges ask whether the path the agent \emph{took} was plausible, whereas \benchmark asks whether the \emph{content} of the final answer is actually present in what the agent \emph{saw} on screen. A trajectory may look plausible, yet end in an unverified claim, and a reasonable-looking answer may have been reconstructed from prior knowledge rather than from the page.

Yet trajectory quality and answer grounding are distinct. A trajectory may visit a relevant page without observing the specific evidence required for the final answer; conversely, an agent may observe relevant information but fail to connect it to its final claims. Prior work has also reported investigative hallucination, where agents claim to have discovered information they never accessed~\citep{pathways}. \benchmark targets this gap by evaluating whether agents observed the required evidence and whether the final answer is supported by it.

\subsection{Observable Evidence Grounding}

Benchmarks on underspecified or ambiguous user requests often focus on whether agents can infer, clarify, or align with user intent~\citep{ambibench,smanbench}. We study a complementary failure mode: an agent may appear to complete the task while silently bridging the gap between user intent and website information using internal knowledge rather than observable evidence.

We formalize this missing dimension as \emph{observable evidence grounding}. The key question is not only whether the final answer is useful, but whether it can be traced back to evidence actually observed during navigation.  To operationalize this, \benchmark pairs real-world tasks with human-curated evidence checks and separates surface completion from evidence observation and claim-level evidence support. This lets us reveal hidden failures that outcome-oriented, static grounding, or trajectory-level evaluations might count as successful.
\section{\benchmark: Evidence-Driven Evaluation of GUI Agents}
\label{sec:evaluation_framework}

\begin{figure*}[t]
    \centering
    \includegraphics[width=\textwidth]{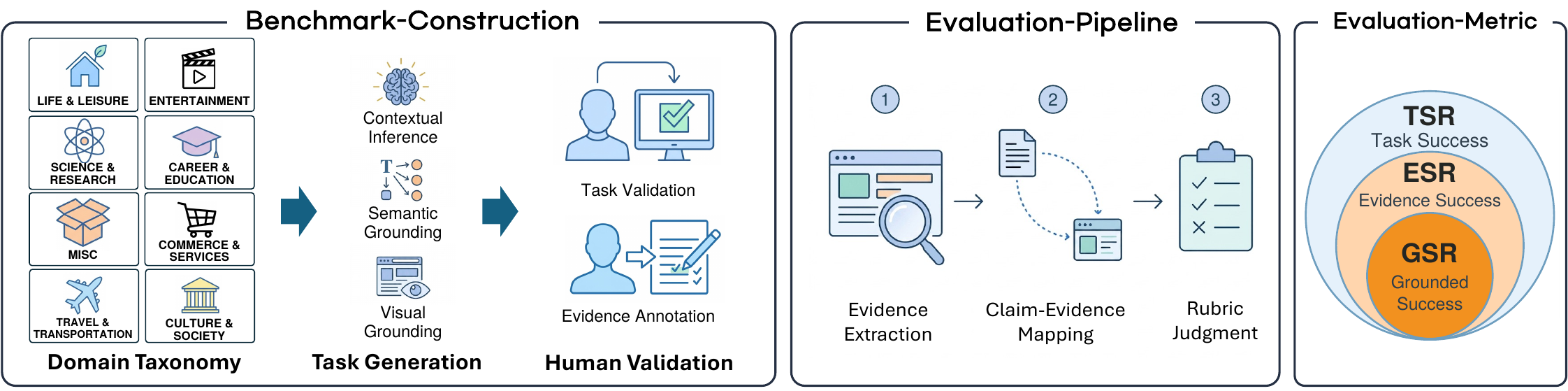}
    \caption{Overview of the EVIDENT task-construction and three-step evidence-aware evaluation pipeline.
    }
\label{fig:pipeline}
\vspace{-0.2in}
\end{figure*}

\subsection{Overview}

\benchmark is designed to evaluate whether a GUI agent's final answer is supported by evidence actually observed during web navigation. Given a user query, trajectory, and a final answer, our goal is not only to determine whether the task appears completed, but also whether the answer can be traced back to evidence available on the visited screens.

This requires addressing two challenges. First, real-world user requests often contain a gap between what the user says and what the website explicitly provides. Second, the evaluation must determine whether the agent observed the evidence needed to bridge this gap and whether its final claims are supported by that evidence. To address the first challenge, we organize tasks by the type of user-intent--website-evidence gap they require. To address the second, we pair each task with human-curated evidence metadata and evaluate trajectories with an evidence-aware judge pipeline. We describe the main design choices in this section and provide the full task construction protocol, annotation guidelines, and prompts in Appendix~\ref{app:benchmark_construction}--\ref{app:judge_pipeline}.

\subsection{Task Taxonomy}

We construct tasks around three recurring gaps between user intent and website information.

\paragraph{Semantic Grounding.}
The user's everyday expression does not directly match the website's UI labels or domain-specific terminology. The agent must map the user's intent to the appropriate site-specific category, filter, label, or textual cue.

\paragraph{Contextual Inference.}
Contextual Inference requires identifying constraints implied by the user's situation. The agent must infer these constraints and verify them on the website, beyond matching the user's wording to website terminology. For example, a request for ``compact furniture for occasional guests'' (Appendix~\ref{sec:task_examples}) implicitly requires identifying a piece that doubles as a sleeping option without dominating a small room.

\paragraph{Visual Grounding.}
The target cannot be identified solely through explicit text. The agent must use visual cues such as icons, badges, images, colors, spatial layout, or product appearance to select or justify the answer.

These categories capture three common forms of evidence gap in real-world tasks: semantic mismatch, implicit constraint inference, and non-textual visual evidence.

\subsection{Human-Curated Evidence Metadata}
\label{sec:annotation_consistency}

\begin{figure}[]
    \centering
    \includegraphics[width=\columnwidth]{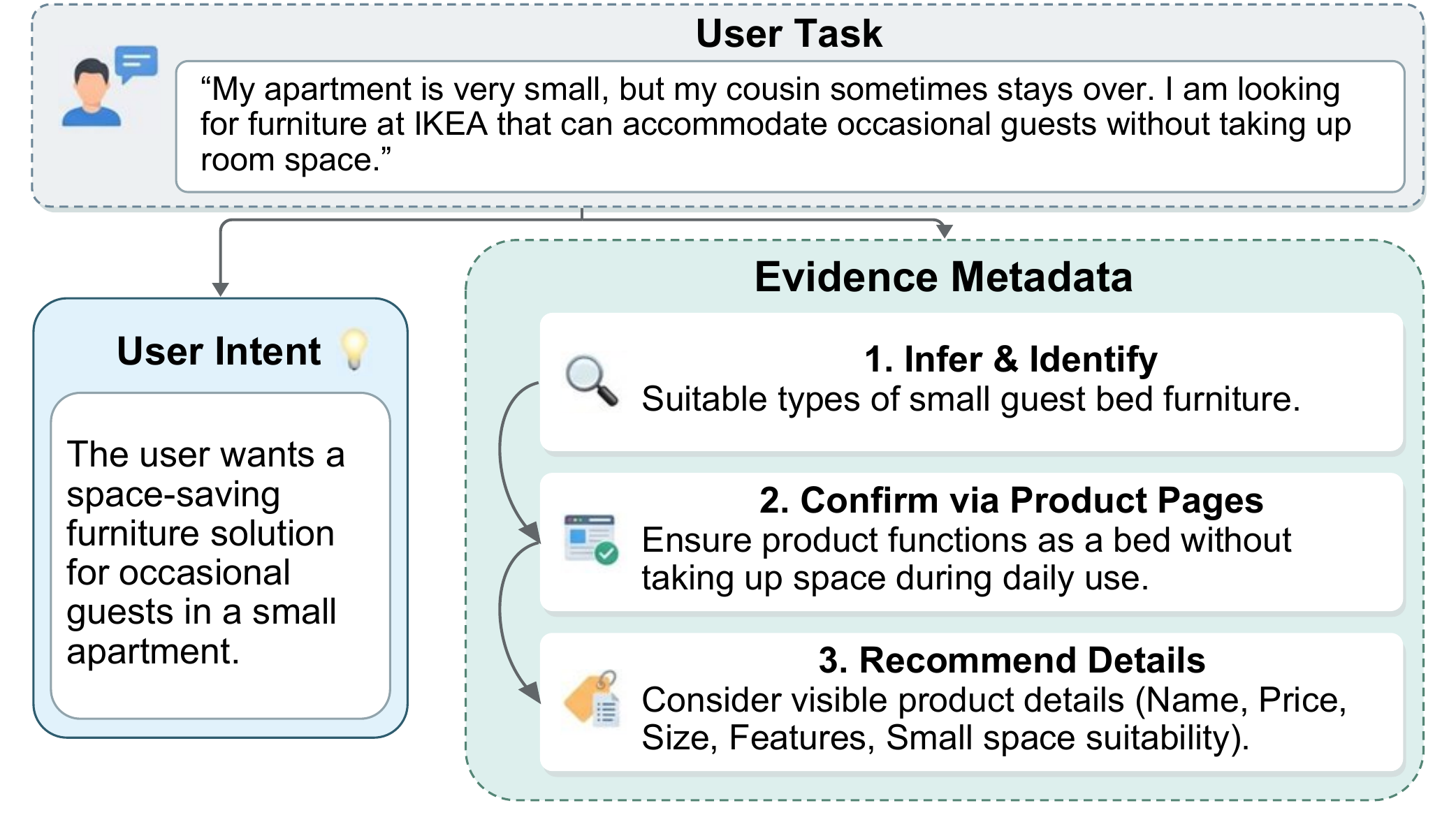}
    \caption{Example of human-curated evidence metadata linking a user request to its intent and required evidence checks.}
    \label{fig:metadata}
    \vspace{-0.2in}
\end{figure}

To evaluate evidence grounding, we must know what evidence an agent should have checked. For each task, we therefore provide human-curated evidence metadata: a set of evidence checks specifying the observable information required to justify a reliable answer.

Evidence metadata specifies what the agent must observe to support its answer. For example, a request for compact furniture for occasional guests may require evidence that an item can serve as a sleeping option, fits a small space, and has suitable dimensions. Different items can satisfy these requirements, provided that the agent observes the relevant website evidence. Appendix~\ref{app:benchmark_construction} describes the annotation criteria, examples, and review procedure.

\paragraph{Checking annotation consistency.}
The evidence requirements should reflect what is needed to answer the user's request, rather than an individual annotator's preferences. To assess their reproducibility, we ask two annotators to independently write evidence-check sets for 50 randomly sampled tasks without seeing the original metadata. A third annotator compares the sets for semantic equivalence. Agreement is assessed over complete requirement sets, rather than individual checks. The independently authored requirements agree on \textbf{48 of 50 tasks (96\%)}. This result shows that different annotators can reproduce the evidence requirements from the task descriptions and shared guidelines.

To assess whether the metadata satisfies the annotation principles, the third annotator examines the existing checks for the same 50 tasks. Each requirement must be observable in the website evidence, necessary to justify an answer, and independent of a particular answer or navigation path. The metadata satisfies these criteria on \textbf{49 of 50 tasks (98\%)}. Together with the agreement between independent annotations, this result supports the use of human-authored evidence checks as reproducible criteria underlying ESR and GSR. Section~\ref{sec:evaluation_validation} separately assesses human--judge agreement, and Appendix~\ref{app:metadata_validation} details the annotation procedure.

\paragraph{Alternative evidence paths.}
The annotation criteria distinguish the information needed to answer a request from the actions used to obtain it. For a request about a protection plan, checks can require the plan's name, price, accident coverage, and warranty terms without prescribing a particular page or sequence of clicks. Different plans and navigation routes can satisfy these requirements. The checks specify the information required to support an answer while allowing different navigation paths. A recommendation can differ from an annotator's preferred choice and still satisfy the requirements when the observed website evidence supports the user's constraints.

\subsection{Evidence-Aware Evaluation}

Directly judging long multimodal trajectories is difficult: trajectories can contain many screenshots, actions, and intermediate observations, and asking a judge model to reason over all of them at once can introduce omissions or hallucinated evidence. We therefore use a three-step evidence-aware evaluation pipeline.

First, the judge extracts observable evidence from trajectory screenshots, including visible text and relevant non-textual cues. Second, it maps both the agent's final-answer claims and the task's evidence metadata to the extracted evidence items. Third, it verifies whether the mapped evidence explicitly supports the required checks and final claims. The evidence checks specify information that must be verified on the website. The judge therefore uses evidence from the trajectory screenshots, rather than parametric knowledge, to determine whether these checks are satisfied (see Appendix~\ref{app:walkthrough_coursera} for an example). This pipeline separates three questions often conflated in outcome-based evaluation: whether the answer appears useful, whether the necessary evidence was observed, and whether the final claims are supported by it.

\paragraph{On the strictness of evidence grounding.}
Evidence grounding measures verification separately from answer correctness. A correct but unverified answer can achieve task success without evidence success or grounded success. Our tasks require time-sensitive or website-specific information, including prices, availability, and visual attributes, which parametric knowledge alone cannot verify at execution time. ESR and GSR measure whether the required evidence is observed and supports the answer.

\subsection{Metrics}

For each task $t \in T$ with user query $q_t$, agent trajectory $\tau_t$, and final answer $y_t$, we define three binary indicators: $s(t) = \mathbb{I}[\mathrm{success}(q_t, y_t)]$ for outcome-level task completion; $e(t) = \mathbb{I}[\mathrm{evidence\_metadata\_satisfied}(\tau_t)]$ for whether the required evidence metadata was observed in the trajectory; and $g(t) = \mathbb{I}[\mathrm{claims\_grounded}(y_t, \tau_t)]$ for whether the final answer's core claims are supported by the observed evidence. We report three success metrics, computed as averages over $T$:
\[
\begin{aligned}
\mathrm{TSR}  &= \frac{1}{|T|}\!\sum_{t} s(t), \\
\mathrm{ESR} &= \frac{1}{|T|}\!\sum_{t} s(t)\cdot e(t), \\
\mathrm{GSR} &= \frac{1}{|T|}\!\sum_{t} s(t)\cdot e(t)\cdot g(t).
\end{aligned}
\]
Task success requires outcome-level completion alone. Evidence success additionally requires observing the required evidence, and grounded success further requires grounding the final claims in that evidence. We report the corresponding rates as \textbf{Task Success Rate (TSR)}, \textbf{Evidence Success Rate (ESR)}, and \textbf{Grounded Success Rate (GSR)}. By construction, $\mathrm{GSR} \leq \mathrm{ESR} \leq \mathrm{TSR}$: the gap $\mathrm{TSR}-\mathrm{ESR}$ captures successful-looking answers produced without observing the necessary evidence, while $\mathrm{ESR}-\mathrm{GSR}$ captures cases where the agent observes the required evidence, but its final answer includes claims that go beyond what that evidence shows. We additionally report average trajectory length in steps as an auxiliary efficiency measure for successful trajectories; its formal definition and implementation details appear in Appendix~\ref{app:metric_details}.

\section{Experiments}\label{sec:exp}

\subsection{Experimental Setup}

\begin{table*}[t]
\centering
\scriptsize
\resizebox{\textwidth}{!}{
\begin{tabular}{llrrrrrr}
\hline
\textbf{Task Type}
& \textbf{Model}
& \multicolumn{3}{c}{\textbf{English}}
& \multicolumn{3}{c}{\textbf{Korean}} \\
\cline{3-8}
&
& \textbf{TSR} & \textbf{ESR} & \textbf{GSR} 
& \textbf{TSR} & \textbf{ESR} & \textbf{GSR} \\
\hline

\multirow{6}{*}{Overall}
& Gemini CUA
& \textbf{87.14} & \textbf{70.00} & \textbf{55.71}
& \textbf{81.33} & \textbf{70.67} & \textbf{45.33} \\
& Claude CUA
& 80.00 & 57.14 & 35.71
& 74.00 & 42.00 & 30.00 \\
& GPT CUA
& 70.00 & 61.43 & 51.43
& 56.67 & 44.00 & 27.33 \\
& MolmoWeb-8B-Native
& 57.14 & 51.43 & 45.71
& 16.00 & 10.00 & 4.67 \\
& Fara-7B
& 57.14 & 50.00 & 35.71
& 8.00 & 0.67 & 0.00 \\
& UI-TARS-1.5-7B
& 54.29 & 35.71 & 22.86
& 24.00 & 13.33 & 10.00 \\
\hline

\multirow{6}{*}{Contextual Inference}
& Gemini CUA
& \textbf{80.95} & \textbf{66.67} & 42.86
& \textbf{84.00} & \textbf{76.00} & \textbf{42.00} \\
& Claude CUA
& \textbf{80.95} & 42.86 & 23.81
& 66.00 & 30.00 & 12.00 \\
& GPT CUA
& 66.67 & 61.90 & \textbf{52.38}
& 44.00 & 30.00 & 16.00 \\
& MolmoWeb-8B-Native
& 57.14 & 47.62 & 42.86
& 16.00 & 10.00 & 6.00 \\
& Fara-7B
& 57.14 & 47.62 & 28.57
& 12.00 & 0.00 & 0.00 \\
& UI-TARS-1.5-7B
& 52.38 & 14.29 & 14.29
& 28.00 & 20.00 & 10.00 \\
\hline

\multirow{6}{*}{Semantic Grounding}
& Gemini CUA
& \textbf{96.30} & \textbf{85.19} & \textbf{81.48}
& 74.00 & \textbf{62.00} & \textbf{54.00} \\
& Claude CUA
& 77.78 & 55.56 & 44.44
& \textbf{78.00} & 56.00 & 50.00 \\
& GPT CUA
& 62.96 & 51.85 & 40.74
& 62.00 & 54.00 & 46.00 \\
& MolmoWeb-8B-Native
& 59.26 & 51.85 & 44.44
& 20.00 & 10.00 & 4.00 \\
& Fara-7B
& 48.15 & 37.04 & 37.04
& 6.00 & 0.00 & 0.00 \\
& UI-TARS-1.5-7B
& 55.56 & 44.44 & 25.93
& 16.00 & 6.00 & 6.00 \\
\hline

\multirow{6}{*}{Visual Grounding}
& Gemini CUA
& \textbf{81.82} & 54.55 & 36.36
& \textbf{86.00} & \textbf{74.00} & \textbf{40.00} \\
& Claude CUA
& \textbf{81.82} & \textbf{72.73} & 36.36
& 78.00 & 40.00 & 28.00 \\
& GPT CUA
& \textbf{81.82} & \textbf{72.73} & \textbf{63.64}
& 64.00 & 48.00 & 20.00 \\
& MolmoWeb-8B-Native
& 54.55 & 54.55 & 50.00
& 12.00 & 10.00 & 4.00 \\
& Fara-7B
& 68.18 & 68.18 & 40.91
& 6.00 & 2.00 & 0.00 \\
& UI-TARS-1.5-7B
& 54.55 & 45.45 & 27.27
& 28.00 & 14.00 & 14.00 \\

\hline
\end{tabular}
}
\caption{Evaluation results across GUI agent models in English and Korean settings. TSR measures outcome success, ESR measures process-level evidence grounding, and GSR measures final-answer grounding.}
\label{tab:combined_results_en_ko}

\end{table*}

\paragraph{Benchmark scope.}
The benchmark contains 220 tasks across eight domains and 31 subcategories, with 70 English and 150 Korean tasks. The language subsets use different websites and are not a matched test of language ability. We report them separately to examine whether evidence-grounding failures recur across these settings.

We test whether \benchmark reveals hidden failures that outcome-oriented evaluation would overlook. We compare closed-source and open-source GUI agents in a unified browser pipeline. Each agent starts from a specified URL, autonomously interacts with the target website, and produces a final answer to the user query. For each task, we log the trajectory as a sequence of screenshots, actions, and intermediate observations. We cap execution at 50 interaction steps to prevent infinite loops.

Completed trajectories are scored using the evidence-aware framework defined in Section~\ref{sec:evaluation_framework}. The judge does not revisit the live website; it only uses screenshots captured during the original agent trajectory as static evidence. This design avoids inconsistencies from dynamic web content and ensures that all judgments are based on the evidence actually available to the agent at execution time. We use Qwen3-VL-235B as the main judge.

\paragraph{Reproducibility.}
Recorded screenshots and actions preserve the evidence available in each trajectory, allowing judgments to be inspected and recomputed without revisiting the site.

\subsection{Agreement with Human Judgments}
\label{sec:evaluation_validation}

\begin{table}[t]
\centering
\small
\begin{tabular}{lcc}
\toprule
Judge setting & Acc. (\%) & $\kappa$ \\
\midrule
Qwen3-VL-235B & 91.27 & 0.864 \\
Avg. over 3 judges & 90.48 & 0.851 \\
\bottomrule
\end{tabular}
\caption{Agreement with human labels on 126 stratified trajectories. Full breakdowns by judge, language, and task type are provided in Appendix~\ref{app:human_agreement}.}
\label{tab:human_agreement_main}
\vspace{-0.1in}
\end{table}

To validate the reliability of this automatic evaluation, we compare judge labels against human annotations on a stratified subset of 126 trajectories covering both languages and all three task types. As shown in Table~\ref{tab:human_agreement_main}, the main judge reaches 91.27\% agreement with human labels ($\kappa=0.864$), and the average over three judge models is 90.48\% ($\kappa=0.851$). Most disagreements occur at the evidence success and grounded success criteria for visual grounding, while outcome-level decisions are nearly always consistent. Additional implementation details, model configurations, and judge-robustness analyses are provided in Appendix~\ref{app:experimental_details} and ~\ref{app:judge_validation}.

\paragraph{Evidence extraction.}
Our extraction step improves the judge's agreement with human judgments compared with using screenshots directly. With the judge model and evaluated trajectories held fixed, agreement ranges from 89\% to 97\% with extracted evidence, compared with 58\% to 72\% with raw-image inputs (Table~\ref{tab:visual_extraction_validation}). The improvement holds across all trajectory-length groups, supporting our choice to extract evidence before judging agent trajectories. Appendix~\ref{app:visual_extraction} describes the input settings.

\begin{table}[t]
\centering
\small
\setlength{\tabcolsep}{1.7pt}
\begin{tabular}{lccc}
\toprule
& \multicolumn{3}{c}{\textbf{Trajectory length (steps)}} \\
\cmidrule(lr){2-4}
\textbf{Judge Input} & \textbf{1--17} & \textbf{18--34} & \textbf{35--50} \\
\midrule
Extracted evidence text (ours) & 89\% & 97\% & 89\% \\
Full trajectory (downsampled) & 72\% & 64\% & 68\% \\
Image windows (8) + prior OCR & 70\% & 58\% & 58\% \\
Image windows (8) only & 63\% & 58\% & 58\% \\
\bottomrule
\end{tabular}
\caption{Agreement with human judgments (\%) by trajectory length.}
\label{tab:visual_extraction_validation}
\vspace{-12pt}
\end{table}

\paragraph{Sources of disagreement.}
Appendix~\ref{app:error_analysis} distinguishes extraction errors from differences in evidence interpretation. In Figure~\ref{fig:case2}, the product photo supports the answer, but its visual details are absent from the extracted catalog. Later stages therefore lack the information needed to verify those claims. In Figure~\ref{fig:case3}, the human annotator and GPT accept the answer based on search results, while Qwen and Gemini require the city's alerts page. The first error concerns what the extractor records, while the second concerns what evidence the judge accepts.

\subsection{Main Results: Surface Success Overestimates Reliability}

Table~\ref{tab:combined_results_en_ko} reports TSR, ESR, and GSR across English and Korean. Commercial agents substantially outperform open-source agents in both languages. Gemini CUA achieves the highest GSR in both English (55.71\%) and Korean (45.33\%).

However, the more important pattern is the consistent gap between TSR and the evidence-grounded metrics. For example, GPT CUA achieves an English TSR of 70.00\%, but its GSR drops to 51.43\% once we require the final claims to be supported by observed evidence. Gemini CUA shows an even larger gap: its English TSR reaches 87.14\%, whereas its GSR is only 55.71\%. In other words, many answers that appear successful under outcome-level evaluation no longer count as successful when we require evidence observation and claim-level grounding.

\begingroup
\widowpenalty=10000
These results support \benchmark's central motivation. Outcome-level success can measure whether an agent produces a useful-looking final answer, but it cannot determine whether the answer is justified by evidence observed on the website. TSR should therefore be interpreted as an upper bound on reliable performance, while ESR and GSR expose hidden failures that are collapsed into success by conventional evaluation.
\par\endgroup

\subsection{Capability Breakdown: Hypothesis Formation vs. Evidence Verification}

\begin{table}[t]
\centering
\scriptsize
\setlength{\tabcolsep}{8pt}
\begin{tabular}{lrrrr}
\toprule
\multirow{2}{*}{\textbf{Dimension}}
& \multicolumn{2}{c}{\textbf{Closed-source}}
& \multicolumn{2}{c}{\textbf{OSS}} \\
\cmidrule(lr){2-3}
\cmidrule(lr){4-5}
& \textbf{En} & \textbf{Ko}
& \textbf{En} & \textbf{Ko} \\
\midrule
Intent Grounding & \textbf{91.2} & \textbf{80.8} & 65.5 & 32.5 \\
Evidence Grounding & 52.5 & 35.2 & 37.9 & 10.7 \\
Response Completion & 81.4 & 70.6 & 57.6 & 16.1 \\
UI label mapping & 87.7 & 75.2 & 67.5 & 36.8 \\
Contextual Inference & 85.2 & 78.0 & \textbf{72.5} & \textbf{37.5} \\
Visual Grounding & 85.8 & 62.1 & 68.2 & 18.1 \\
\bottomrule
\end{tabular}
\caption{Average rubric dimension pass rates (\%) by model group and language. Closed-source: Gemini, Claude, GPT; OSS: Molmo, FARA, UI-TARS.}
\label{tab:rubric_dimension_group_results}
\vspace{-0.2in}
\end{table}

To identify where agents struggle, we use an additional rubric covering six diagnostic dimensions and compare the scores across model groups and languages (Table~\ref{tab:rubric_dimension_group_results}). This analysis examines where GUI agents fail when moving from understanding user intent to evidence-grounded verification.

Across both model groups and languages, Intent Grounding pass rates far exceed Evidence Grounding pass rates. This gap suggests that current GUI agents are relatively better at forming plausible hypotheses from user requests than at validating those hypotheses against live website evidence. An agent may understand what it should look for, but still fail to check the concrete prices, schedules, specifications, badges, images, or other conditions needed to justify its final answer.

Contextual Inference further reveals differences across languages and model groups. Commercial agents score 85.2\% in English and 78.0\% in Korean on Contextual Inference, compared with 72.5\% and 37.5\%, respectively, for open-source agents. These results indicate that open-source agents remain particularly limited on the Korean tasks, which require inferring implicit user constraints and connecting them to site-specific evidence. Details are given in Appendix~\ref{app:c2}.

\subsection{Is Evidence Grounding Merely Longer Navigation?}

We finally test whether evidence-grounded success is simply a byproduct of longer navigation. We compare average trajectory length across three nested groups: trajectories achieving task success, evidence success, and grounded success. TSR, ESR, and GSR report the proportions of trajectories achieving each type of success, respectively. Full results are provided in Appendix~\ref{app:trajectory_statistics}. For example, Gemini CUA in Korean averages 18.87 steps for trajectories with task success but 15.94 steps for trajectories with grounded success; GPT CUA shows the same pattern, dropping from 27.49 to 23.71 steps. In English, Gemini CUA also remains comparable, with 13.89 steps for trajectories with task success and 13.38 for trajectories with grounded success. Trajectories with grounded success do not consistently require more steps than trajectories with task success.

\section{Analysis}\label{sec:analysis}

Section~\ref{sec:exp} showed that outcome-level success can overestimate GUI-agent reliability. We now analyze what this gap contains: which hidden failures are collapsed into surface success and where they emerge most frequently.

\subsection{Decomposing Surface Success}
\label{subsec:error_analysis}

\begin{table}[!h]
\centering
\scriptsize
\setlength{\tabcolsep}{5pt}
\renewcommand{\arraystretch}{1.15}
\begin{tabular}{@{}p{2.8cm} p{4.1cm}@{}}
\toprule
\textbf{State} & \textbf{Description} \\
\midrule
\multicolumn{2}{@{}c}{\textbf{Task success}} \\
\midrule
\textit{True Success} 
& Observes the required evidence and grounds its final claims in it. \\
\textit{Claim-Ungrounded} \newline \textit{Success} 
& Observes relevant evidence but the final answer is not sufficiently supported. \\
\textit{Hallucinated Success} 
& Answer appears successful despite missing the required evidence. \\
\midrule
\multicolumn{2}{@{}c}{\textbf{No task success}} \\
\midrule
\textit{Task Failure}
& Fails to meet the task-success criterion. \\
\bottomrule
\end{tabular}
\caption{Four evaluation states based on task success, evidence observation, and claim grounding.}
\label{tab:evaluation_states}
\vspace{-0.15in}
\end{table}

We classify trajectories into the four mutually exclusive states in Table~\ref{tab:evaluation_states} using task success, evidence observation, and claim grounding. Trajectories that do not complete the task are classified as \textit{Task Failure}.

\begin{figure}[!t]
    \centering
    \includegraphics[width=\columnwidth]{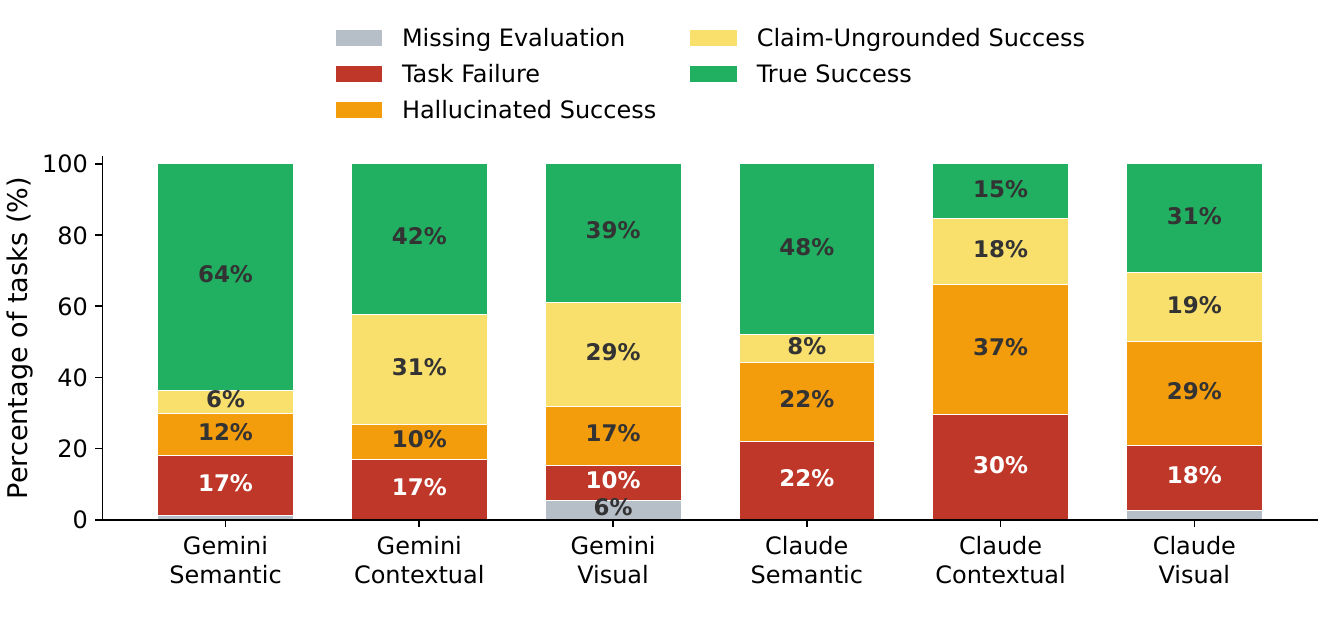}
    \vspace{-0.3in}
    \caption{Error pattern composition across task types.}
    \label{fig:error_composition}
\vspace{-0.1in}
\end{figure}

Figure~\ref{fig:error_composition} reports the distribution of these states across task categories, using the criteria defined above. \textit{Hallucinated Success} achieves task success but not evidence success: the answer looks complete, but the agent never observed the supporting evidence. \textit{Claim-Ungrounded Success} achieves evidence success but not grounded success: the agent observes the required evidence, but its final answer includes claims that go beyond what that evidence shows. Grounded success corresponds to the \textit{True Success} label.

Both models reach their highest \textit{True Success} on Semantic Grounding (64\% Gemini, 48\% Claude). The two models also fail in different ways on harder categories: Gemini's \textit{Claim-Ungrounded Success} reaches 31\% on Contextual Inference and 29\% on Visual Grounding, meaning it observes the evidence but does not ground its answer in it. Figure~\ref{fig:error_composition} shows that 37\% of Claude's Contextual Inference trajectories and 29\% of its Visual Grounding trajectories are classified as \textit{Hallucinated Success}, achieving task success but not evidence success. Evidence-aware metrics therefore tell us not just how often agents fail, but how.

\subsection{Hidden Failures Behind Outcome-Level Success}
  
\begin{figure}[t]
    \centering
    \includegraphics[width=0.88\columnwidth]{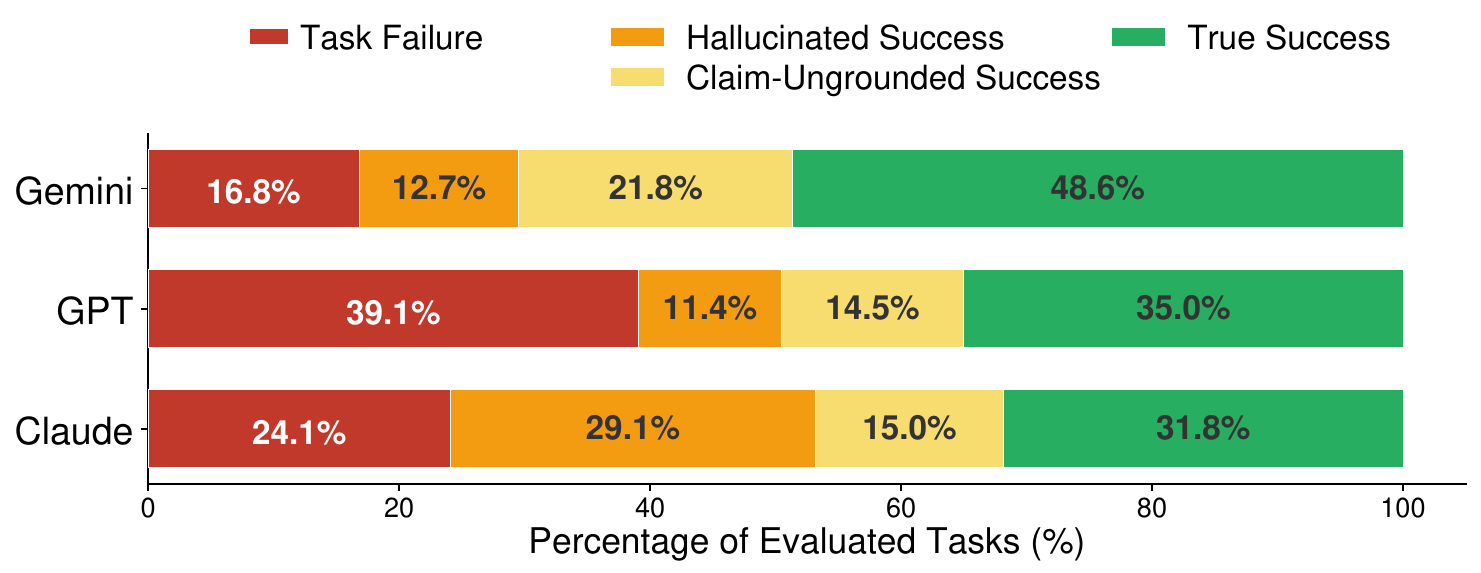}
    \vspace{-1em}
    \caption{Error composition across commercial models.}
    \label{fig:error_composition_all}
    \vspace{-1.5em}
\end{figure}

Figure~\ref{fig:error_composition_all} shows the distribution of the four evaluation outcomes for the commercial models. This decomposition separates genuinely grounded answers from plausible but unsupported ones. Across commercial agents, \textit{Hallucinated Success} and \textit{Claim-Ungrounded Success} account for a substantial share of trajectories with task success. For Gemini CUA, the strongest model in our evaluation, TSR reaches 83.2\%, but this aggregate includes both grounded and ungrounded answers. Specifically, only 48.6\% result in \textit{True Success}, while 34.5\% result in hidden failures: 21.8\% are \textit{Claim-Ungrounded Success} and 12.7\% are \textit{Hallucinated Success}.

This result directly supports the motivation of \benchmark. A high outcome-level success rate alone does not establish reliable web-based decision-making. Many successful-looking answers are plausible because they are consistent with the model's prior knowledge or general assumptions, but they are not traceable to evidence observed during the trajectory. We provide a detailed example in the Coursera example shown in Appendix~\ref{app:walkthrough_coursera}. The agent remains on the landing page without searching for or opening the relevant program page, yet still recommends Google Data Analytics and justifies it as beginner-friendly from prior knowledge. Such cases are risky in live web environments because the agent never verified website-specific evidence.

\paragraph{Human validation of hidden failures.}
Three annotators independently review all 27 Gemini trajectories labeled \textit{Hallucinated Success} by the Qwen judge, using the request, full screenshot-and-action trajectory, final answer, and evidence checks. All three annotators identify missing required evidence in 25 of 27 trajectories (92.6\%), agreeing with the judge's Hallucinated Success label.

\subsection{Where Do Hidden Failures Emerge?}
\label{subsec:taxonomy_analysis}

We next analyze which task conditions most strongly expose the gap between surface success and evidence-grounded success, focusing on task type and difficulty.

\begin{figure}[!t]
    \vspace{-0.1in}
    \centering
    \includegraphics[width=\columnwidth]{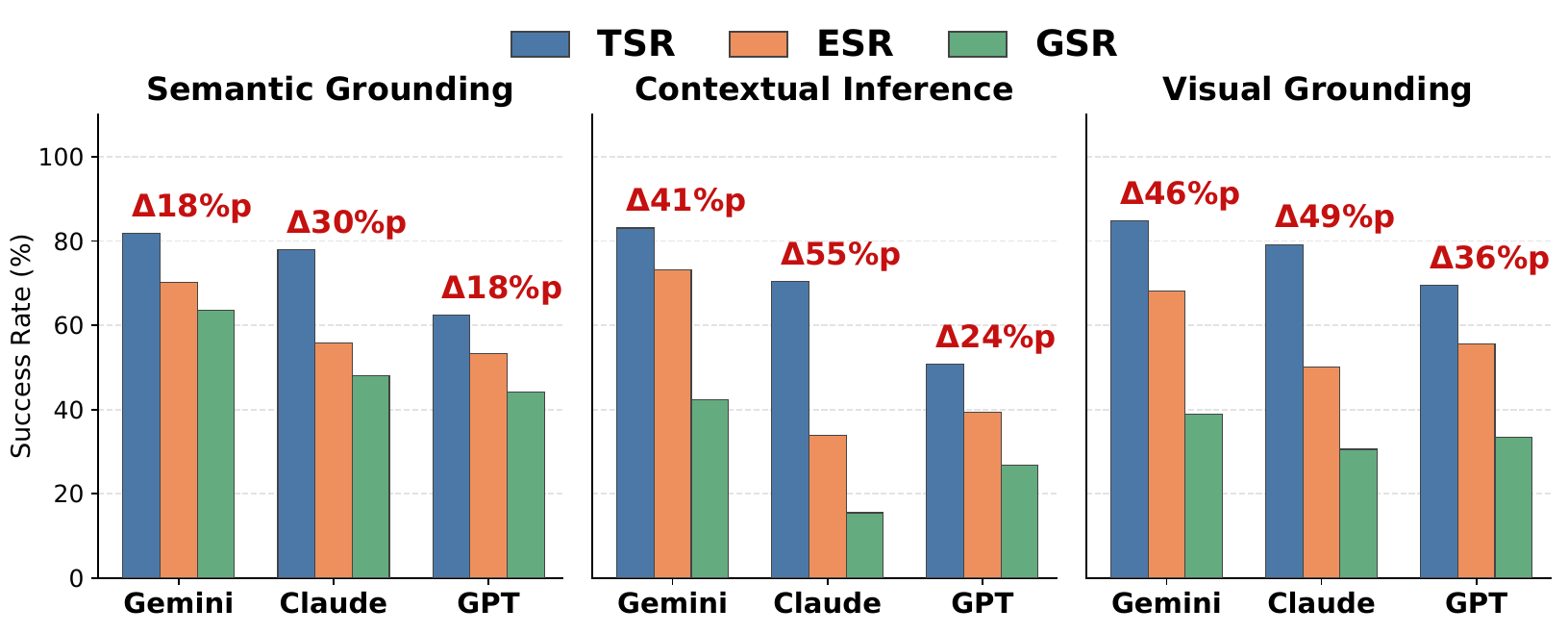}
    \vspace{-0.3in}
    \caption{Average performance gap between TSR and GSR by task type.}
    \label{fig:gaps_by_task_type}
\vspace{-0.2in}
\end{figure}

\paragraph{By task type.}
As shown in Figure~\ref{fig:gaps_by_task_type}, Semantic Grounding tasks yield the highest average GSR and the narrowest gap between task completion and grounded success. These tasks mainly require mapping user expressions to explicit UI vocabulary, which current agents handle relatively well when the relevant labels are visible. In contrast, the gap becomes much larger for Contextual Inference and Visual Grounding. The average GSR drops significantly for these categories, even though their TSR remains substantially higher.

Hidden failures are most likely when agents must go beyond surface keyword matching. Contextual tasks require inferring implicit constraints and verifying them against site-specific information, while visual tasks require non-textual cues such as icons, badges, colors, images, or layout. The association between task type and failure mode is statistically significant ($\chi^2=20.63$, $p<0.01$).%

\begin{figure}[!t]
    \vspace{-0.1in}
    \centering
    \includegraphics[width=0.73\linewidth]{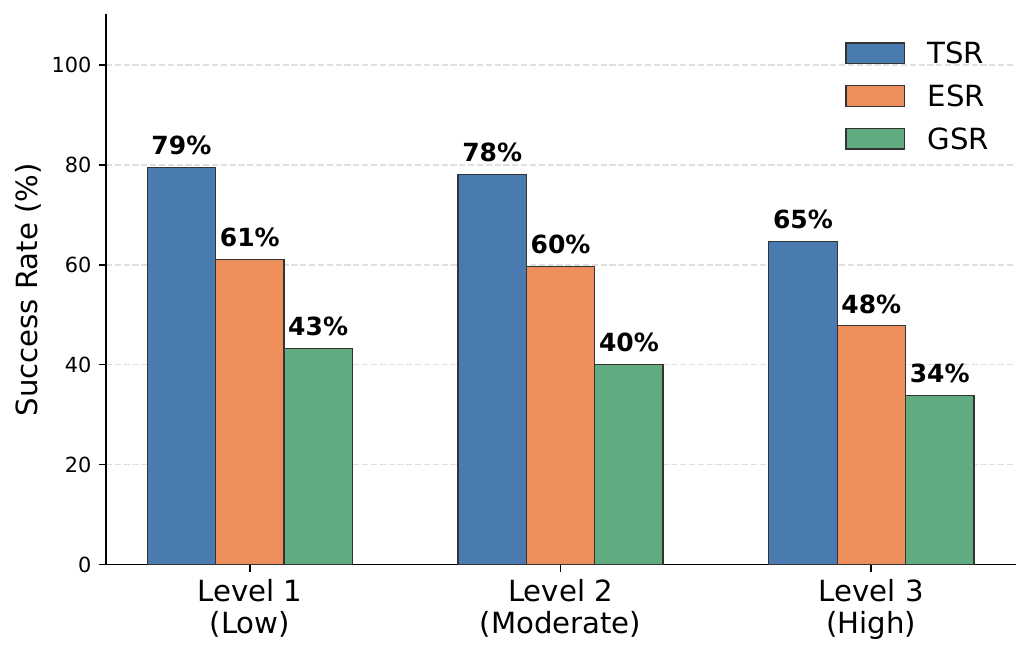}
    \vspace{-0.1in}
    \caption{Performance (TSR, ESR, GSR) decreases monotonically as task difficulty increases.}    \label{fig:difficulty_performance}
\vspace{-0.2in}
\end{figure}

\paragraph{By difficulty.}

We also group tasks by difficulty using \textit{Constraint Depth}, the number of required evidence conditions, and \textit{Information Density}, the normalized instruction length; see Appendix~\ref{sec:appendix_difficulty}. As shown in Figure~\ref{fig:difficulty_performance}, all three metrics decrease monotonically with difficulty. Specifically, GSR drops from 43.2\% at Level 1 to 33.8\% at Level 3. Higher difficulty is associated with lower TSR and GSR.

\section{Conclusion}

We introduced \benchmark, an evidence-driven benchmark for evaluating whether GUI agents ground their final answers in evidence they observe during web navigation. Unlike outcome-oriented evaluation, \benchmark separates surface task completion from evidence observation and claim-level grounding through TSR, ESR, and GSR. Our experiments show that current GUI agents often achieve high surface success while failing to verify the evidence needed to support their answers. This gap is especially large in contextual and visual grounding tasks, where agents must infer implicit constraints or use non-textual cues. These findings motivate improvements in evidence collection and claim verification alongside navigation.

\section*{Limitations}

This work has several limitations. First, \benchmark covers English and Korean web tasks, with a larger Korean subset. This design allows us to stress-test evidence grounding in a non-English web environment that is less represented in existing GUI-agent benchmarks, but it should not be interpreted as a population-level comparison between English and Korean web use. The two subsets may differ in website conventions, layout density, and task ecology. We therefore report results separately by language and treat cross-language differences as diagnostic rather than as controlled estimates of language difficulty.

Second, \benchmark is built on real-world websites. This improves ecological validity, but it also limits reproducibility because website content, layouts, prices, schedules, and availability can change over time. Our evaluation addresses this by judging agents only against screenshots captured during their original trajectories, rather than revisiting the live website. As a result, \benchmark evaluates whether an agent grounded its answer in the evidence it actually observed at execution time, but it does not provide a permanently fixed snapshot of each website.

Third, the evidence metadata is human-curated. This is necessary for evaluating whether an answer depends on task-relevant evidence rather than merely matching a final string, but it also means that the benchmark reflects annotator decisions about which evidence checks are necessary. Some tasks may allow alternative valid evidence paths that are not fully captured by the current metadata. Future versions could improve coverage by collecting multiple independent annotations or explicitly marking optional evidence checks.

Fourth, our evaluation relies on an LLM-based judge for long multimodal trajectories. We reduce this risk through a structured extraction--mapping--judgment pipeline and validate the judge against human annotations, but automatic judging can still miss subtle visual cues, extract incomplete evidence, or draw boundaries differently from humans when deciding whether a claim is explicitly supported. This limitation is especially relevant for visual grounding tasks, where the relevant evidence may be carried by images, icons, or layout rather than text.

Finally, \benchmark adopts a conservative definition of observable evidence grounding. This choice is intentional: the benchmark is designed to reveal cases where agents answer plausibly without verifying the website evidence their answers depend on. However, this strict criterion may under-credit answers that combine observed evidence with reasonable commonsense inference. The resulting scores should therefore be read as diagnostic measures of evidence-grounded behavior, not as a complete measure of user satisfaction or practical task usefulness.

\bibliography{custom}
\clearpage
\appendix
\makeatletter
\setlength{\@fptop}{0pt}
\setlength{\@dblfptop}{0pt}
\makeatother

\section{Computer-Use Agents (CUAs)}
\label{Appendix:Computer-Use-Agents}
\label{app:experimental_details}

\subsection{CUA}
A Computer-Use Agent (CUA) is a multimodal agent that manipulates Graphical User Interfaces (GUIs), such as web browsers or operating-system screens, based on natural language instructions from the user.
A CUA observes the current screen, selects the next action to take toward a given goal, and then receives the updated screen as input to determine its subsequent action.
In other words, a CUA accomplishes tasks through a repeated interaction loop of observing the screen, generating an action, executing it in the environment, and updating the state.

A general vision-language model (VLM) can accept images or screens as input and produce descriptions, answer questions, perform OCR, and perform visual reasoning.
However, to be used in a Web GUI environment, the model must also be capable of outputting actions that can be executed in a browser.
Therefore, a model used in a Web GUI CUA must be able to recognize elements within the screen, bridge user instructions with the current screen state, and generate actions, such as clicking, typing, scrolling, navigating to a page, selecting options, and producing a final answer, in a predefined format.

The CUA system in this work consists of an execution framework that connects the model to the browser environment.
At each step, the system provides the model with the user query, a screenshot of the current screen, a textual observation, and the history of previous actions.
The model outputs the next action to perform, and the system applies it to a Playwright-based browser environment.
The system then records the updated screenshot and metadata, which are used as the observation for the next step.
This process repeats until the model produces a final answer or the task meets its termination condition.

In this work, all models are evaluated using the same CUA execution pipeline based on MolmoWeb.
Every model runs under identical settings for the browser environment, observation format, action schema, trajectory logging pipeline, and termination rules.
The only adjustments per model are the tokenizer, image preprocessing, and output parsing, adapted to match each model's input and output format.

\subsection{Model Descriptions}

All API-based models were evaluated in May 2026 using the model identifiers exposed by each provider at the time of evaluation. We used deterministic decoding settings where the API exposes decoding controls. Across all models, we kept the same browser environment, observation format, action schema, trajectory logging pipeline, and 50-step cap.

\paragraph{Gemini CUA.}
For the Gemini-based CUA, this work uses \texttt{gemini-3-flash-preview}.
This proprietary multimodal model is served via API and takes user instructions and web screen observations as input to predict the next action.

\paragraph{GPT CUA.}
For the OpenAI-based CUA, this work uses \texttt{gpt-5.4}.
This proprietary multimodal model can process both text and image inputs; here, we use it to generate the next action based on web screen observations.

\paragraph{Claude CUA.}
For the Claude-based CUA, this work uses \texttt{claude-opus-4-7}.
This is a proprietary API model supporting long-context reasoning and multimodal input processing, used here to predict the next action from user instructions and web screen observations.

\paragraph{microsoft/Fara-7B.}
\href{https://huggingface.co/microsoft/Fara-7B}{\texttt{microsoft/Fara-7B}} is a 7B-scale open-weight computer-use agent released by Microsoft.
The model is designed to visually observe web pages and generate GUI actions such as clicking, typing, and scrolling.

\paragraph{ByteDance-Seed/UI-TARS-1.5-7B.}
\href{https://huggingface.co/ByteDance-Seed/UI-TARS-1.5-7B}{\texttt{ByteDance-Seed/UI-TARS-1.5-7B}} is a 7B-scale open-weight GUI agent model from the UI-TARS family.
It is built around visual grounding and user interface action prediction.

\paragraph{allenai/MolmoWeb-8B-Native.}
\href{https://huggingface.co/allenai/MolmoWeb-8B-Native}{\texttt{allenai/MolmoWeb-8B-Native}} is an 8B-scale open-weight multimodal web agent from Ai2's MolmoWeb series.
The model targets browser-based interaction and generates actions for completing web tasks from visual observations.

\clearpage\section{Details of Task Collection}

\subsection{Domain Distribution}
\begin{figure*}[!t]
    \centering
    \includegraphics[width=0.65\textwidth]{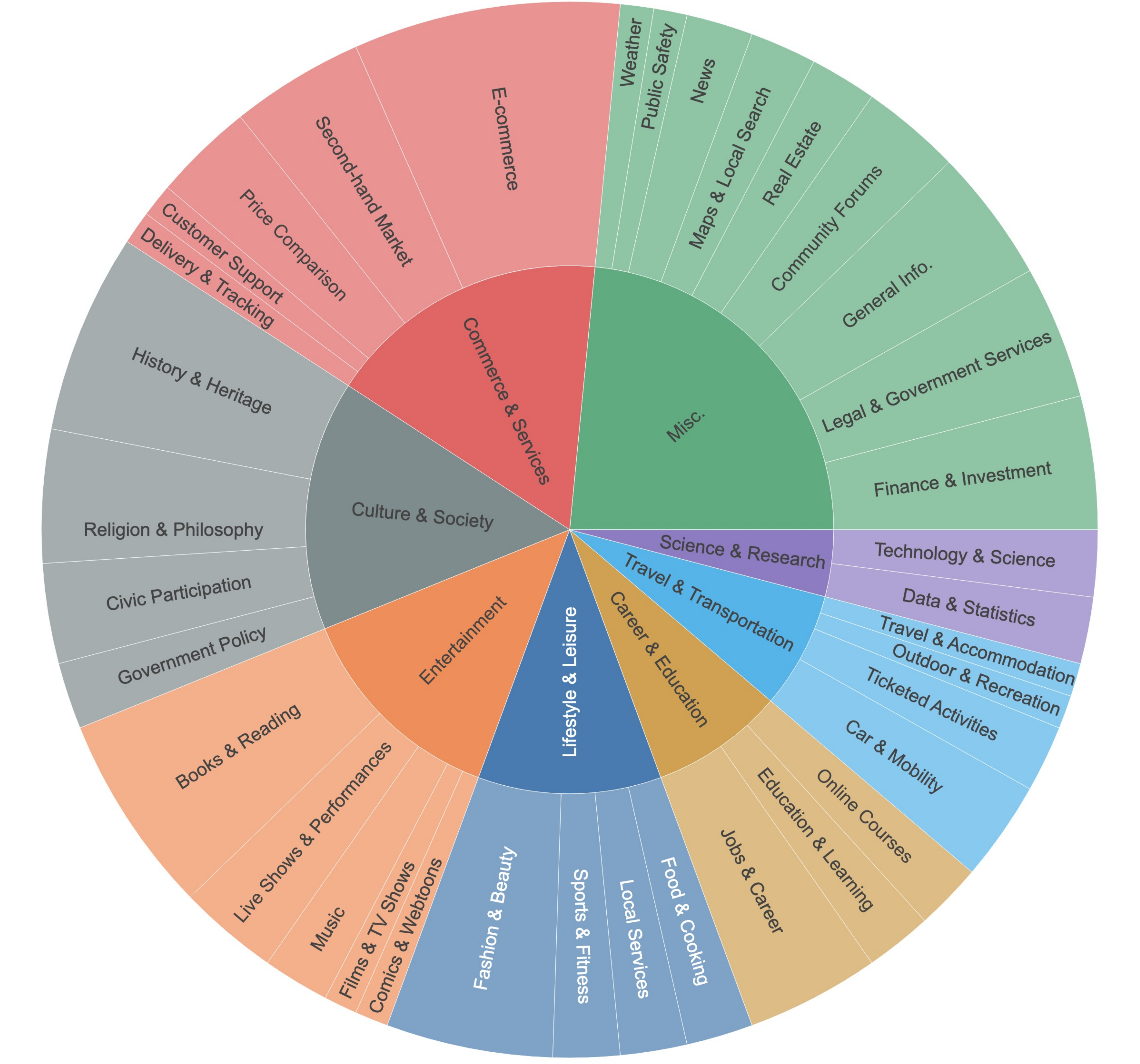}
    \vspace{-0.1in}
    \caption{Website categories in the benchmark. The inner ring shows primary categories and the outer ring shows subcategories.}
    \label{fig:domain_sunburst}
    \vspace{-0.1in}
\end{figure*}

Figure~\ref{fig:domain_sunburst} presents the hierarchical domain distribution of our evaluation dataset. The dataset covers 8 primary categories: Lifestyle \& Leisure, Entertainment, Misc., Science \& Research, Career \& Education, Travel \& Transportation, Commerce \& Services, and Culture \& Society, which are further divided into 31 subcategories. The websites selected span highly popular and domain-specific real-world web applications. The angular size of each sector reflects the frequency of target websites mapped to each taxonomy, showing high domain diversity and the authentic web scenarios computer-use agents must navigate.

\subsection{Task and Difficulty Distributions}

Figure~\ref{fig:dataset_difficulty_distribution} shows the distribution of task difficulty levels across English (70 tasks) and Korean (150 tasks). The language split is intentionally asymmetric. We oversample Korean to stress-test evidence grounding in a less-studied non-English web environment, where task construction often involved dense layouts, visual cues, and locale-specific website conventions. The English subset serves as a complementary reference setting rather than an equally sized control group. Accordingly, we report results separately by language and do not interpret aggregate scores as population-level language comparisons.

Figure~\ref{fig:dataset_task_type_distribution} shows the task-type distribution across Semantic Grounding, Contextual Inference, and Visual Grounding. The dataset maintains an approximately balanced division among the three categories (77 Semantic, 71 Contextual, and 72 Visual tasks), enabling comparison across different forms of user-intent--website-evidence gaps.

\subsection{Design Principles of Tasks}
The tasks in our dataset were constructed following two core design principles to evaluate the execution fidelity of computer-use agents (CUAs):
\begin{enumerate}
    \item \textbf{Process-Level Constraint Verification:} Rather than relying on simple end-state string matches, each task defines multiple explicit evidence checks (as detailed in Appendix~\ref{sec:appendix_difficulty}) that require the agent to visually inspect, confirm, and interact with specific UI components.
    \item \textbf{Cognitive and Contextual Demand:} We explicitly designed tasks requiring contextual inference (e.g., matching a user's subtle mood preference to a hotel's review sentiment or photo vibe) rather than simple keyword matches.
\end{enumerate}
\begin{figure}[!t]
\begin{minipage}[t]{0.48\textwidth}
\input{Figures/fig_dataset_difficulty_distribution}
\end{minipage}\hfill
\begin{minipage}[t]{0.48\textwidth}
\input{Figures/fig_dataset_task_type_distribution}
\end{minipage}
\end{figure}

\raggedbottom
\section{Ablation Study}

\begin{table}[h]
\centering
\scriptsize
\setlength{\tabcolsep}{3pt}
\resizebox{\columnwidth}{!}{
\begin{tabular}{lrrrrrr}
\toprule
\multirow{2}{*}{\textbf{Setting}}
& \multicolumn{3}{c}{\textbf{English}}
& \multicolumn{3}{c}{\textbf{Korean}} \\
\cmidrule(lr){2-4}
\cmidrule(lr){5-7}
& \textbf{TSR}
& \textbf{ESR}
& \textbf{GSR}
& \textbf{TSR}
& \textbf{ESR}
& \textbf{GSR} \\
\midrule
No constraint
& 87.14 & 70.00 & \textbf{55.71}
& 81.33 & 70.67 & \textbf{45.33} \\

No internal knowledge
& 71.43 & 64.29 & 48.57
& 59.33 & 52.67 & 41.33 \\

\quad + few-shot
& 61.43 & 57.14 & \textbf{45.71}
& 51.33 & 44.00 & \textbf{33.33} \\
\bottomrule
\end{tabular}
}
\caption{Gemini results under different prompt settings.}
\label{tab:gemini_prompt_setting_results}
\end{table}
Table~\ref{tab:gemini_prompt_setting_results} compares the Gemini-based agent's default baseline (``No constraint'') with two system-prompt variants: ``No internal knowledge'' and ``No internal knowledge + few-shot''.

All three success metrics decrease as the prompt constraints become stricter. The ``No internal knowledge'' condition reduces English GSR from 55.71 to 48.57 and Korean GSR from 45.33 to 41.33. Adding few-shot examples of evidence-based navigation further reduces GSR to 45.71 in English and 33.33 in Korean.

These results show that prompt constraints alone are insufficient to improve evidence grounding.

\begingroup
\interlinepenalty=10000
\subsection{Evaluation across Judge Models}
\label{app:judge_validation}
To evaluate consistency across judge models, we scored the same agent trajectories with \texttt{gpt-5.4-mini} and \texttt{gemini-3.1-flash-lite} in addition to the primary judge, Qwen3-VL-235B (Figure~\ref{fig:judge_llm_en_ko}).

All three judges consistently assign higher average TSR, ESR, and GSR scores to commercial models than to open-source models across both languages and all task types.

Absolute scores, gap sizes, and individual judgments differ across judges, especially for visual evidence (Appendix~\ref{app:error_analysis}).
\begin{figure}[!t]
    \centering
    \includegraphics[width=1.0\columnwidth]{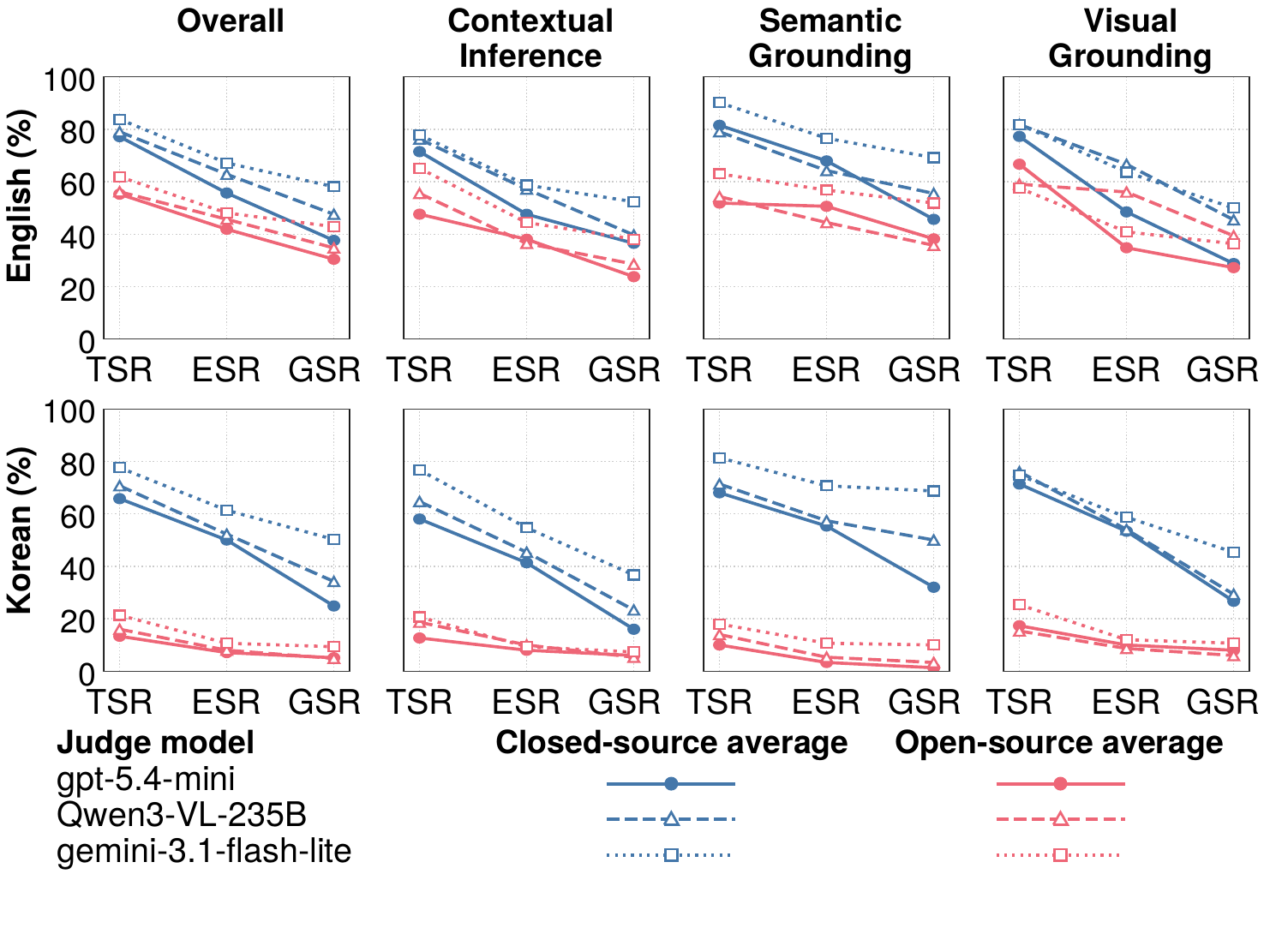}
    \caption{Evaluation results across three judge models.}
    \label{fig:judge_llm_en_ko}
\end{figure}

\subsection{Understanding Requests and Verifying Evidence}
\label{app:c2}

The diagnostic rubric examines how agents interpret requests and use website evidence. Intent Grounding, Evidence Grounding, and Response Completion are diagnostic dimensions assessed across all tasks. User Interface Label Mapping, Contextual Inference, and Visual Grounding are task-specific diagnostic dimensions. Table~\ref{tab_rubric_all} defines the underlying criteria, and Table~\ref{tab:rubric_dimension_results} reports scores by model and language.
\begin{table*}[!t]
\centering
\small
\setlength{\tabcolsep}{4pt}
\begin{tabular*}{\textwidth}{@{\extracolsep{\fill}}llrrrrrr@{}}
\toprule
\multirow{2}{*}{\textbf{Model}} & \multirow{2}{*}{\textbf{Lang.}}
& \multicolumn{3}{c}{\textbf{All tasks}} & \multicolumn{3}{c}{\textbf{Task-specific}} \\
\cmidrule(lr){3-5}\cmidrule(lr){6-8}
& & \shortstack{Intent\\Grounding} & \shortstack{Evidence\\Grounding} & \shortstack{Response\\Completion}
& \shortstack{User Interface\\Label Mapping} & \shortstack{Contextual\\Inference} & \shortstack{Visual\\Grounding} \\
\midrule
\multirow{2}{*}{Gemini} & En & \textbf{94.1} & 55.9 & \textbf{89.7} & \textbf{100.0} & 84.1 & 80.0 \\
& Ko & \textbf{90.5} & \textbf{44.2} & \textbf{82.3} & \textbf{79.6} & \textbf{92.0} & \textbf{82.3} \\
\addlinespace[3pt]
\multirow{2}{*}{Claude} & En & 91.2 & 36.8 & 82.4 & 77.8 & \textbf{95.2} & \textbf{90.0} \\
& Ko & 80.0 & 28.7 & 72.7 & 70.0 & 74.7 & 51.0 \\
\addlinespace[3pt]
\multirow{2}{*}{GPT} & En & 88.2 & \textbf{64.7} & 72.1 & 85.2 & 76.2 & 87.5 \\
& Ko & 72.0 & 32.7 & 56.7 & 76.0 & 67.3 & 53.0 \\
\addlinespace[3pt]
\multirow{2}{*}{Molmo} & En & 64.7 & 50.0 & 58.8 & 59.3 & 71.4 & 62.5 \\
& Ko & 36.7 & 12.9 & 16.3 & 43.8 & 44.7 & 18.4 \\
\addlinespace[3pt]
\multirow{2}{*}{FARA} & En & 64.2 & 37.3 & 59.7 & 77.8 & 81.2 & 76.3 \\
& Ko & 17.6 & 2.0 & 8.1 & 30.6 & 20.4 & 4.0 \\
\addlinespace[3pt]
\multirow{2}{*}{UI-TARS} & En & 67.6 & 26.5 & 54.4 & 65.4 & 65.0 & 65.9 \\
& Ko & 43.3 & 17.3 & 24.0 & 36.0 & 47.3 & 32.0 \\
\bottomrule
\end{tabular*}
\par\smallskip
\begin{minipage}{\textwidth}
\footnotesize
En: English; Ko: Korean.\\
For Contextual Inference and Visual Grounding, scores average the percentages of trajectories satisfying each criterion listed in Table~\ref{tab_rubric_all}.
\end{minipage}
\caption{Rubric scores (\%) by model and language.}
\label{tab:rubric_dimension_results}
\end{table*}

\paragraph{Intent Grounding without Evidence Grounding}
The results show a severe performance gap between Intent Grounding and Evidence Grounding, observed consistently across all models and language settings. In English, Gemini and Claude score \textbf{94.1\%} and \textbf{91.2\%} on Intent Grounding but only \textbf{55.9\%} and \textbf{36.8\%} on Evidence Grounding, respectively, showing a gap between understanding the user's goal and supporting the selected target and reasons with observed website evidence.

This pattern is even more pronounced in the Korean setting and among open-source (OSS) models. For example, FARA not only struggled to understand the user's intent in Korean, achieving only \textbf{17.6\%} on Intent Grounding, but also showed an almost complete failure in verifying concrete facts such as numerical values or specifications from the visible web page, scoring \textbf{2.0\%} on Evidence Grounding. These results demonstrate that current GUI agents can often formulate plausible task interpretations by relying on the strong linguistic understanding of LLMs, but frequently skip the crucial Evidence Grounding step of checking whether text and visual evidence from the live website support those interpretations.

\paragraph{Inferring requirements from user context}
Contextual Inference assesses whether agents go beyond understanding the user's stated goal to identify requirements implied by the user's situation. Table~\ref{tab:rubric_dimension_results} shows that Gemini and Claude achieve high Contextual Inference scores in both languages, while all three open-source models score lower in Korean than in English. This language difference indicates that the open-source models infer contextual requirements less consistently on the Korean tasks.

\subsection{Evaluation on an External Benchmark}
\begin{figure}[!t]
    \centering
    \includegraphics[width=0.68\columnwidth]{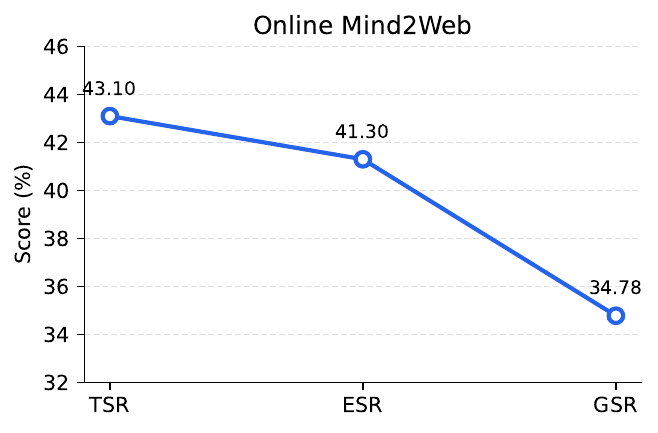}
    \caption{Evaluation results on the Online Mind2Web benchmark.}
    \label{fig:online_mind2web}
\end{figure}

To examine whether our findings extend beyond the tasks curated for \benchmark, we evaluate 45 randomly sampled tasks from Online Mind2Web. Human annotators mapped essential visual cues according to our evidence metadata guidelines.

Figure~\ref{fig:online_mind2web} reports a TSR of \textbf{43.10\%} and an ESR of \textbf{41.30\%}. Because most Online Mind2Web requests specify a particular destination or action, agents often successfully navigate to the final destination (TSR) by following text-based cues, but frequently skip the crucial intermediate evidence pages or detailed information windows (ESR) along the way.

Requiring the final answer to be grounded in the observed evidence reduces the success rate from \textbf{41.30\%} (ESR) to \textbf{34.78\%} (GSR). This gap represents \textit{Claim-Ungrounded Success}: the trajectory contains the required evidence, but the final answer includes claims that go beyond what that evidence shows. Figure~\ref{fig:case1} illustrates this distinction in an EVIDENT task. The agent observes the recreation area's photos but claims they show a child playing in the sand, a detail the human annotators do not support.

These findings highlight a critical flaw in conventional metrics that merely assess whether the agent reaches the target page and show why evaluating trustworthiness requires assessing task completion separately from visual evidence verification using human-defined criteria.

\endgroup
\subsection{Multi-Seed Comparison on Open-Source Models} 
\begin{figure}[htbp]
    \centering
    \includegraphics[width=\columnwidth]{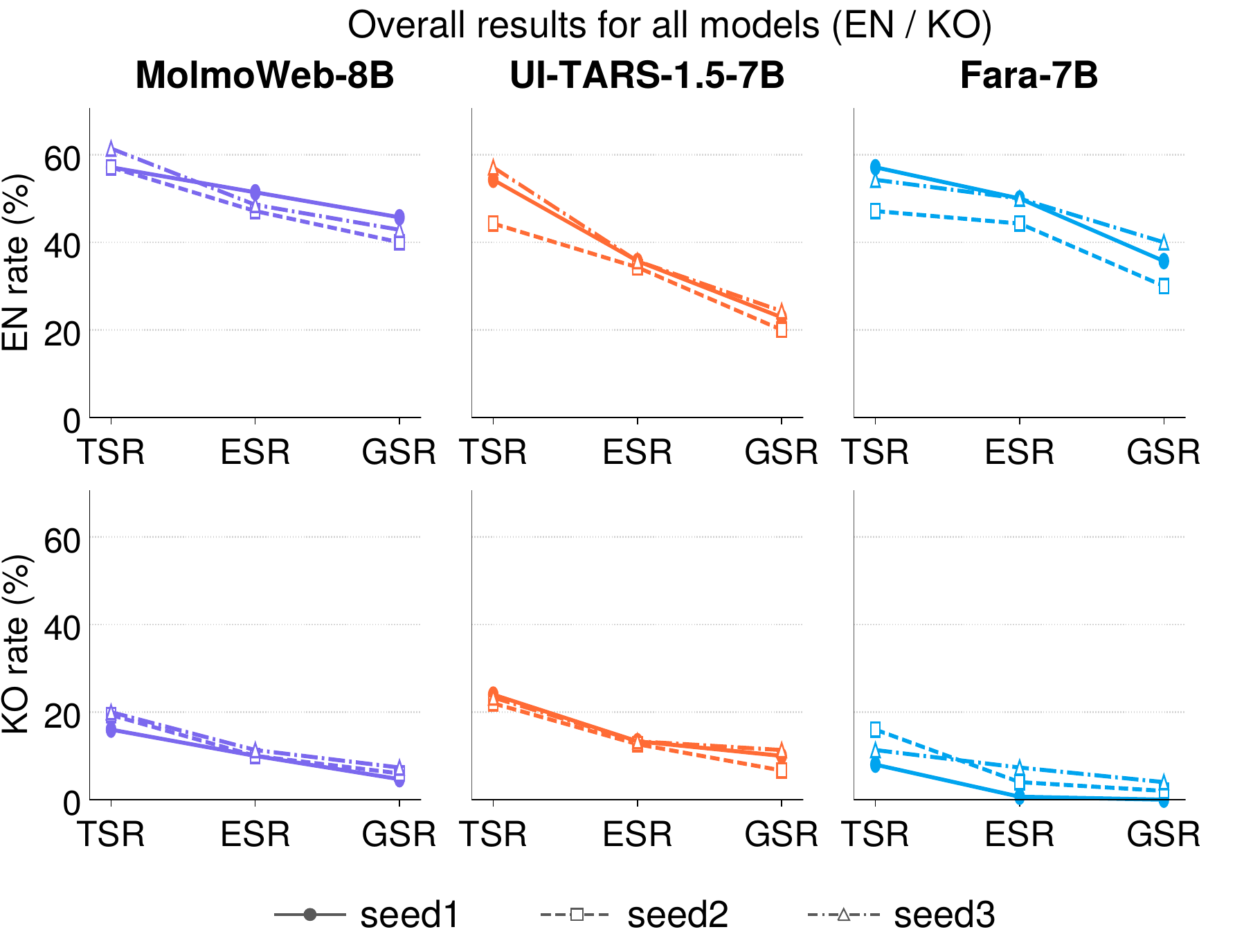}
    \caption{Multi-seed robustness results. Gaps between TSR, ESR, and GSR recur across seeds.}
    \label{fig:multiseed}
\vspace{-0.2in}
\end{figure}
To assess the robustness of our findings, we evaluate the open-source models using multiple random seeds.

Figure~\ref{fig:multiseed} shows that absolute scores vary slightly across seeds, while the proportions of \textit{Hallucinated Success} and \textit{Claim-Ungrounded Success} remain similar for each model.

\begingroup
\interlinepenalty=10000
\setlength{\intextsep}{2pt}
\setlength{\textfloatsep}{5pt}
\setlength{\floatsep}{5pt}
\setlength{\abovecaptionskip}{3pt}
\subsection{Trajectory Statistics}\label{app:trajectory_statistics}
We obtain valid evaluations for 1,299 trajectories across six models on the 70 English and 150 Korean tasks.

Figure~\ref{fig:trajectory_step_distribution} displays the distribution of trajectory lengths (step counts) across the models. Claude CUA averages 12.0 steps in Korean and 16.0 steps in English. In contrast, GPT CUA and microsoft/Fara-7B average 35.8 steps and 32.9 steps, respectively, in Korean. GPT CUA and Fara-7B reach the 50-step limit more frequently than the other models, especially on Korean tasks.

\begin{figure}[htbp]
    \centering
    \includegraphics[width=\columnwidth]{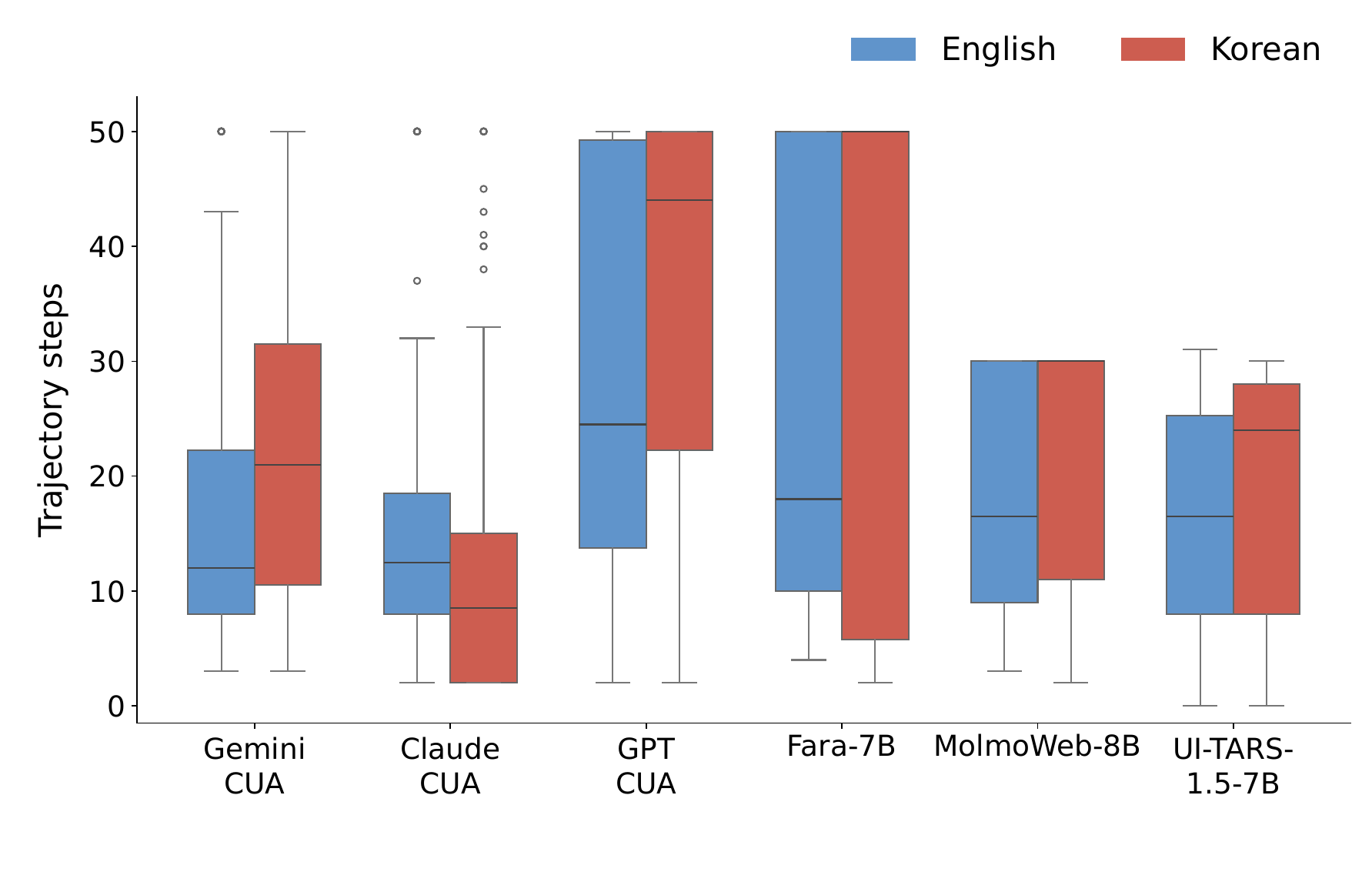}
    \caption{Trajectory lengths for the 1,299 evaluated trajectories.}
    \label{fig:trajectory_step_distribution}
\end{figure}

\begin{table}[!htbp]
\centering
\scriptsize
\begin{tabular}{llrrr}
\toprule
\textbf{Lang.}
& \textbf{Model}
& \textbf{TSR}
& \textbf{ESR}
& \textbf{GSR} \\
\midrule
\multirow{6}{*}{English}
& Gemini CUA     & 13.89 & 14.06 & 13.38 \\
& Claude CUA     & 13.66 & 14.62 & 15.16 \\
& GPT CUA        & 23.73 & 24.58 & 24.11 \\
& MolmoWeb-8B    & 13.70 & \textbf{13.31} & 12.72 \\
& Fara-7B        & 13.75 & 13.91 & \textbf{12.08} \\
& UI-TARS-1.5-7B & \textbf{12.97} & 13.32 & 13.31 \\
\midrule
\multirow{6}{*}{Korean}
& Gemini CUA     & 18.87 & 17.45 & 15.94 \\
& Claude CUA     & 12.75 & 18.02 & 17.36 \\
& GPT CUA        & 27.49 & 28.38 & 23.71 \\
& MolmoWeb-8B    & 12.29 & 11.87 & 13.71 \\
& Fara-7B        & \textbf{12.25} & \textbf{2.00}  & --    \\
& UI-TARS-1.5-7B & \textbf{12.25} & 13.75 & \textbf{13.13} \\
\bottomrule
\end{tabular}
\par\smallskip
\begin{minipage}{\columnwidth}
\footnotesize
-- indicates no trajectories with grounded success.
\end{minipage}
\caption{Average trajectory length (steps) by success criterion.}
\label{tab:ats_results_model_lang}

\end{table}

Table~\ref{tab:ats_results_model_lang} reports average trajectory length in steps for three nested groups of trajectories: those with task success, evidence success, and grounded success. These groups are used to compute TSR, ESR, and GSR, respectively. The averages are computed separately for each group and show that trajectories with grounded success are not consistently longer than those in the other two groups. Longer browsing alone does not account for evidence-grounded success.

\endgroup
\clearpage
\section{Additional Details of \benchmark Construction}
\label{app:benchmark_construction}

Appendix~\ref{app:benchmark_construction} provides additional details on task collection and evidence metadata annotation. The full judge prompts are provided in Appendix~\ref{app:judge_pipeline}.

\subsection{Task Collection and Taxonomy Assignment}

We constructed \benchmark from real-world user scenarios that require agents to bridge a gap between user intent and website information. Each task consists of a user query, a start URL, a task category, and human-curated evidence metadata. We assign each task to one of three categories: Semantic Grounding, Contextual Inference, or Visual Grounding.

Semantic Grounding tasks require mapping everyday user expressions to website-specific labels or terminology. Contextual Inference tasks require inferring implicit constraints from the user context and verifying them against website information. Visual Grounding tasks require identifying targets through non-textual cues such as icons, badges, images, colors, or spatial layout.

When a task involved multiple types of gaps, annotators assigned the primary category according to the evidence most central to justifying the final answer. For example, if a task required both interpreting a user's preference and checking a visual badge, annotators labeled it as Visual Grounding when the visual cue was necessary to select the final target.

\subsection{Evidence Metadata Annotation}
\label{app:evidence_metadata}

For each task, annotators created evidence metadata specifying the evidence that must be observed to justify a reliable answer. Evidence metadata was designed as a checklist of observable facts rather than as a single gold answer. This allows different valid final answers, as long as they are supported by the required evidence.

Each evidence check was written to satisfy three criteria:
\begin{enumerate}
    \item \textbf{Observability}: the evidence must be verifiable from the website screen or trajectory screenshots.
    \item \textbf{Necessity}: the evidence must be necessary to justify the final answer for the user query.
    \item \textbf{Answer-independence}: the evidence check should not force a specific wording of the final answer, but should specify what must be checked.
\end{enumerate}

For example, for a task asking the agent to find compact furniture for occasional guests in a small apartment, the evidence metadata may require checking that the item can serve as a sleeping option, that it is suitable for small-space use, and that visible product details such as dimensions or functional descriptions were inspected.

\begingroup
\interlinepenalty=10000
\subsection{Evidence Metadata Validation}
\label{app:metadata_validation}

\begin{table}[!ht]
\centering
\small
\begin{tabular}{lcc}
\toprule
\textbf{Validation} & \textbf{\#Ann.} & \textbf{Agreement} \\
\midrule
Independent re-authoring & 2 & 48/50 (96\%) \\
Metadata audit & 1 & 49/50 (98\%) \\
\bottomrule
\end{tabular}
\caption{Evidence metadata validation on 50 randomly sampled tasks.}
\label{tab:evidence_validation}
\end{table}

We assess whether the evidence requirements are reproducible across annotators and whether the existing metadata satisfies the annotation principles. The study uses 50 randomly sampled tasks and three annotators. To ensure independent annotation, the two annotators write evidence-check sets from the task descriptions and shared guidelines without access to the original metadata. The third annotator compares their sets for semantic equivalence at the task level. Because the checks are free-form requirements, agreement means that the two sets require equivalent evidence, rather than identical wording or categorical labels. The sets agree on 48 of 50 tasks (96\%), as summarized in Section~\ref{sec:annotation_consistency}.

To assess the existing metadata, the third annotator applies the principles in Appendix~\ref{app:evidence_metadata}: observability, necessity, and answer-independence. Each check must specify information needed to support the answer without imposing a particular answer or navigation path. The metadata satisfies these criteria on 49 of 50 tasks (98\%). Table~\ref{tab:evidence_validation} reports the independent-annotation agreement and metadata-validation results. These findings support the consistency of the evidence requirements across annotators and their adherence to the annotation principles, providing a reliable basis for evaluating evidence grounding.

\endgroup

\clearpage
\begin{table*}[t] 
\centering
\scriptsize
\begin{tcolorbox}[
    title={metadata rules},
    width=\textwidth,
    colback=blue!3,
    colframe=blue!50,
    coltitle=white,
    fonttitle=\bfseries
]
\begin{lstlisting}[basicstyle=\fontsize{7}{7.5}\selectfont\ttfamily, breaklines=false]
## Annotator Guidelines: Writing Evidence Checks for GUI Agents

**Core Principle**
Focus on what information was verified. The evidence_checks define the critical pieces of information that 
must be logically verified from the screenshot OCR data before the final answer.

---

### 1. Flexible Target Identification

Do not force a single rigid UI path. The check should evaluate whether the correct type of information or
candidate was found, regardless of the exact navigation path used.

| | Example |
|---|---|
| Bad | The search bar was used to search for 'sofa bed'. |
| Good | A compact furniture option suitable for sleeping was identified. |

---

### 2. Structured 3-Step Flow

Keep evidence_checks concise, typically around three items, following the logical flow of the task.

| Step | Name | Meaning |
|---|---|---|
| 1 | Intent Mapping | The ambiguous user request was translated into an appropriate constraint. |
| 2 | Target Identification | A valid candidate matching the required category or scenario was identified. |
| 3 | Condition Verification | The domain-relevant conditions necessary for the final judgment - such as
date/time, cost, eligibility, policy, rating, or availability - were verified. |

---

### 3. Indispensability Test

Ask whether the final answer would be a lucky guess if this specific fact had not been verified. If yes,
include it. If the condition is trivial or not necessary for the task goal, leave it out.

> **Example:** If the user asks for yoga clothes with sweat absorption and stretch, then material features
related to sweat absorption and stretch should be included as evidence checks.

---

### JSON Template & Example

{
  "json_template": {
    "id": "task_id_example",
    "prompt": "Actual user utterance written in natural, everyday language.",
    "task_type": "Contextual Inference / Semantic Grounding / Visual Grounding",
    "start_url": "https://www...",
    "user_intent": "A clear, one-sentence summary of the user's actual goal.",
    "evidence_checks": [
      "The ambiguous request was mapped to an appropriate constraint.",
      "A valid candidate fitting the required scenario was identified.",
      "The final answer was supported by the domain-relevant conditions necessary for the final judgment,
       such as date/time, cost, eligibility, policy, rating, or availability."
    ]
  },
  "example": {
		"id": "ikea_small_apartment_guest_sleeping",
		    "prompt": "My apartment is tiny, but my cousin sometimes sleeps over. On IKEA, find a 
                       furniture option that can help with occasional guests without taking over the room.",
		    "task_type": "Contextual Inference",
		    "start_url": "https://www.ikea.com/us/en/",
		    "user_intent": "The user wants a space-saving furniture solution for occasional overnight 
                            guests in a small apartment.",
		    "evidence_": [
		      "A suitable compact guest-sleeping furniture type was inferred and identified.",
		      "The product page or listing confirmed that the item can function as a sleeping option
             without dominating the room during everyday use",
		      "The recommendation considered visible product details such as product name, price,
             dimensions, functional description, or small-space suitability"
		    ]
  }
}
\end{lstlisting}
\end{tcolorbox}
\caption{Annotation guidelines for evidence metadata}
\label{fig_metadata_rules}
\end{table*}
\clearpage
\section{Evidence-Aware Judge Pipeline}
\label{app:judge_pipeline}

The judge model scores a \emph{completed} agent trajectory (it does not run the agent). It proceeds in three steps: extracting observable evidence, mapping the answer's claims and the human evidence checks onto it, and checking evidence support and assigning scores. The pipeline produces TSR, ESR, and GSR together with a final label for each trajectory. We detail each step below and illustrate the evaluation process in Appendix~\ref{app:walkthrough_coursera}.

\subsection{Step 1: Observable Evidence Extraction}

In the first step, the judge model receives trajectory screenshots and extracts observable evidence. The output includes visible text, UI labels, buttons, filters, list items, prices, ratings, dates, and other screen-level information. For visual grounding tasks, the judge also records non-textual cues such as icons, badges, image content, color cues, selected states, and layout positions.

This step is intentionally restricted to evidence extraction. The judge is not allowed to decide task success, compare the trajectory to the final answer, or infer hidden information. Agent thoughts are not treated as admissible evidence unless the corresponding information is visible on the screen. Appendix~\ref{app:walkthrough_coursera} (Step~1) shows the \texttt{evidence\_catalog} this produces for an example trajectory.

\subsection{Step 2: Claim--Evidence and Metadata--Evidence Mapping}

In the second step, the judge receives the extracted evidence items, the agent's final answer, and the task's evidence metadata. It decomposes the final answer into core claims and maps each claim to the evidence items that may support it. It also maps each evidence metadata check to evidence items observed in the trajectory.

\newpage
The output of this step is a candidate alignment between claims, required evidence checks, and observed evidence. This alignment is not treated as the final judgment; it is re-examined in Step 3.

\subsection{Step 3: Rubric Judgment}

In the third step, the judge verifies whether the mapped evidence explicitly supports the required evidence checks and final-answer claims. The judge may only use information contained in the extracted evidence items. It must not rely on commonsense knowledge, assumptions about the website, or the model's parametric knowledge.

The final rubric contains three common dimensions:
\begin{itemize}
    \item \textbf{Intent Grounding}: whether the agent correctly understood the user's core goal.
    \item \textbf{Evidence Grounding}: whether the agent's output is explicitly supported by observed website evidence.
    \item \textbf{Response Completion}: whether the final response completes the user's request.
\end{itemize}

Task-type-specific criteria are additionally applied. Semantic Grounding checks whether the user's expression is mapped to the appropriate UI label or site-specific terminology. Contextual Inference checks whether implicit constraints are correctly inferred and verified. Visual Grounding checks whether non-textual cues are correctly interpreted and used to select or justify the target. The rubric criteria are diagnostic and do not affect the success metrics. Appendix~\ref{app:walkthrough_coursera} applies the three steps to a Coursera task, showing how the evidence checks and claims are evaluated and how the final label is assigned.

\clearpage

\begin{table*}[p]
\centering
\footnotesize
\setlength{\abovecaptionskip}{2pt}
\setlength{\belowcaptionskip}{0pt}

\tcbset{
  rubricbox/.style={
    width=\textwidth,
    boxsep=1pt, left=3pt, right=3pt, top=2pt, bottom=2pt,
    coltitle=white,
    fonttitle=\bfseries\small,
    fontupper=\scriptsize,
    enhanced, sharp corners=south,
    before skip=2pt, after skip=2pt
  }
}

\begin{tcolorbox}[rubricbox, title={Rubric Dimensions: Common},
                  colback=blue!5, colframe=blue!75!black]
\begin{lstlisting}[basicstyle=\scriptsize\ttfamily,
                   breaklines=false, aboveskip=0pt, belowskip=0pt,
                   xleftmargin=0pt, framexleftmargin=0pt]
intent_grounding:
  Intent Grounding -- Did the agent correctly interpret the user's
  request and core goal?

evidence_grounding:
  Evidence Grounding --
  (1) Was the selected or recommended target, such as a product, recipe,
      article, or item, actually visible in the merged evidence through
      screen or text evidence?
      If the target is insufficiently verified or not evidenced, this
      dimension fails.
  (2) Was the reason for selecting that target grounded in site-observed
      facts, such as price, specifications, filters, availability,
      description, ingredients, quantities, rankings, reviews, or visible
      attributes?
      If the reason relies only on general knowledge, assumptions, or
      plausible inference, this dimension fails.
  (Hard check: concrete values such as Korean-won prices, bold product
   names, quantities, ingredients, and named items must match the Step 1
   evidence strings in screenshot_ocr or visual_elements.)

response_completion:
  Response Completion -- Did the final answer fully complete the user's
  request?
\end{lstlisting}
\end{tcolorbox}

\begin{tcolorbox}[rubricbox, title={Rubric Dimensions: Semantic Grounding},
                  colback=teal!5, colframe=teal!75!black]
\begin{lstlisting}[basicstyle=\scriptsize\ttfamily,
                   breaklines=false, aboveskip=0pt, belowskip=0pt,
                   xleftmargin=0pt, framexleftmargin=0pt]
ui_label_mapping:
  UI Label Mapping -- Did the agent correctly map the user's expression
  to the intended UI meaning, such as labels, menus, filters, tags,
  sorting options, or site-specific categories?
  (Refer to task_spec.user_expression and expected_grounding.)
\end{lstlisting}
\end{tcolorbox}

\begin{tcolorbox}[rubricbox, title={Rubric Dimensions: Contextual Inference},
                  colback=orange!5, colframe=orange!75!black]
\begin{lstlisting}[basicstyle=\scriptsize\ttfamily,
                   breaklines=false, aboveskip=0pt, belowskip=0pt,
                   xleftmargin=0pt, framexleftmargin=0pt]
context_to_requirement:
  Context-to-Requirement Extraction -- Did the agent correctly extract
  the actionable requirements needed to complete the task from the
  user's situational context?

implicit_need_validity:
  Implicit Need Validity -- Were the inferred implicit requirements
  reasonable and necessary given the user's situation?

priority_tradeoff:
  Priority & Tradeoff Handling -- When not all conditions could be
  perfectly satisfied, did the agent prioritize the core requirements
  and make a reasonable substitution or tradeoff?
\end{lstlisting}
\end{tcolorbox}

\begin{tcolorbox}[rubricbox, title={Rubric Dimensions: Visual Grounding},
                  colback=purple!5, colframe=purple!75!black]
\begin{lstlisting}[basicstyle=\scriptsize\ttfamily,
                   breaklines=false, aboveskip=0pt, belowskip=0pt,
                   xleftmargin=0pt, framexleftmargin=0pt]
visual_intent_grounding:
  Visual Intent Grounding -- Did the agent interpret the user's
  references to location, icons, badges, colors, images, or other
  visual cues as concrete visual conditions to look for?

visual_target_disambiguation:
  Visual Target Disambiguation -- Did the agent select the correct
  target associated with the visual cue described by the user?
\end{lstlisting}
\end{tcolorbox}

\caption{Common and task-specific rubric dimensions. Common dimensions apply to all tasks. Task-specific dimensions cover Semantic Grounding, Contextual Inference, and Visual Grounding.}
\label{tab_rubric_all}
\end{table*}

\clearpage
\begin{table*}[t] 
\centering
\scriptsize
\begin{tcolorbox}[title={Prompt for step1}, width=\textwidth]
\begin{lstlisting}[basicstyle=\fontsize{7}{7.5}\selectfont\ttfamily, breaklines=false]
You are a **screenshot OCR** pass for a GUI/web agent trajectory.

## Your role
For each attached screenshot, extract visible on-screen text as faithfully as possible.
Your priority is OCR-like text extraction, not image captioning.
Do not write long image descriptions.Do not infer hidden meanings.Do not judge task success.Do not compare to
final_answer.Do not decide whether evidence supports the answer.Agent thought may help locate frames, but 
is not admissible as page evidence.

## Screenshot OCR
For **every attached screenshot**, emit one `screenshot_ocr` row.
Each row should contain:
- readable visible text from the screenshot
- UI labels, menu labels, tabs, buttons, filters, selected options
- card/list item text: titles, subtitles, prices, ratings, counts, badges, dates, rankings
- search query text if visible
- URL bar text only if visible in the screenshot
- uncertainty notes for unreadable or partially readable text
Preserve the original language and numbers.
Preserve important line breaks when helpful.
Do not summarize when exact text is readable.
If text is partially unreadable, write the readable part and mark unreadable spans as `[unclear]`.
Include important visual-only cues inside `ocr_text` when they matter and have no readable label, prefixed 
with `VISUAL:`:
- visible badge/icon without readable text
- product photo contents (box, accessory, logo, color, object)
- selected/disabled/highlighted state, ranking position, or spatial relation

## Non-empty extraction
If any screenshot shows a loaded page, emit at least one `screenshot_ocr` row per screenshot.
`screenshot_ocr: []` is valid only when every image is blank, loading-only, error-only, or login/captcha wall.
When OCR text appears slightly corrupted but the visible screenshot strongly indicates a common Korean
product term, preserve the most visually plausible text and add the uncertainty note.

Examples:
- Do not silently invent; write the corrected text in ocr_text and mention the ambiguity in
unreadable_or_uncertain.

## Visual Element Extraction (for Visual Grounding)
For each screenshot, also emit a `visual_elements` list for non-text visual cues that could serve as selection 
targets.

Each entry:
- `element_type`: "icon" | "badge" | "image" | "color_cue" | "layout_position"
- `description`: what it looks like (e.g. "CARE certification icon, shield-shaped")
- `location`: spatial anchor (e.g. "top-right of the 3rd product card", "left side of the search bar")
- `associated_text`: nearby readable text if any
- `screenshot_index_in_request`: number

## Output
Return ONLY valid JSON:

{
  "window_index": number | null,
  "screenshot_ocr": [
    {
      "screenshot_index_in_request": number,
      "trajectory_step": number | null,
      "screenshot_filename": string | null,
      "ocr_text": string,
      "unreadable_or_uncertain": string
    }
  ],
  "visual_elements": [
    {
      "screenshot_index_in_request": number,
      "element_type": "icon" | "badge" | "image" | "color_cue" | "layout_position",
      "description": string,
      "location": string,
      "associated_text": string | null
    }
  ],
  "window_notes": string
}

Rules:
- `ocr_text` should be OCR-like text, not a prose description; include all readable UI text in one field.
- Do not output checkpoint judgments.
- Do not output task success.
- Do not use agent thought as evidence.
\end{lstlisting}
\end{tcolorbox}
\caption{Evaluation prompt for Step 1.}
\label{fig_prompt_for_step1}
\end{table*}
\clearpage
\begin{table*}[t] 
\centering
\scriptsize
\begin{tcolorbox}[title={Prompt for step2}, width=\textwidth]
\begin{lstlisting}[basicstyle=\fontsize{7}{7.5}\selectfont\ttfamily, breaklines=false]
You are Step 2 evidence mapping for GUI/web trajectory evaluation.

`evidence_catalog` is built from Step 1 `screenshot_ocr` and `visual_elements` rows.
OCR entry ids look like `screen_ocr_1`, `screen_ocr_2`, ...; visual entry ids look like `screen_visual_1`,
`screen_visual_2`, ...
Pick `supporting_evidence_ids` from those ids and quote the matching snippet in `evidence_quote`.

Return ONLY valid JSON with exactly these top-level keys:
- final_answer_claims
- evidence_check_reviews

Critical rule: Evidence fields may be empty, but required rows must never be omitted.

If input.final_answer is non-empty:
- final_answer_claims MUST contain at least one row.
- Even unsupported claims must be included.
- For unsupported claims, use:
  supporting_evidence_ids=[]
  evidence_quote=""
  satisfied=false

If input.eval_checkpoints has N items:
- evidence_check_reviews MUST contain exactly N rows.
- Preserve checkpoint order.
- Copy checkpoint id into checkpoint_id.
- Even unsupported checkpoints must be included.
- For unsupported checkpoints, use:
  supporting_evidence_ids=[]
  evidence_quote=""
  satisfied=false
  notes explaining what is missing.

Use `satisfied`, not `supported` or `verified`.

A visible title/search result supports only existence.
Recommendation reasons, suitability, ease, popularity, taste, healthiness, ingredients, quantities, or methods
require explicit visible evidence.

Required output shape:
{
  "final_answer_claims": [
    {
      "claim": "string",
      "claim_type": "existence",
      "is_core_claim": false,
      "supporting_evidence_ids": [],
      "evidence_quote": "",
      "satisfied": false
    }
  ],
  "evidence_check_reviews": [
    {
      "checkpoint_id": "ec_0",
      "description": "string",
      "supporting_evidence_ids": [],
      "evidence_quote": "",
      "satisfied": false,
      "notes": "No direct visible evidence."
    }
  ]
}
\end{lstlisting}
\end{tcolorbox}
\caption{Evaluation prompt for Step 2.}
\label{fig_prompt_for_step2}
\end{table*}
\clearpage
\begin{table*}[t] 
\centering
\scriptsize
\begin{tcolorbox}[title={Prompt for step3}, width=\textwidth]
\begin{lstlisting}[basicstyle=\fontsize{7}{7.5}\selectfont\ttfamily, breaklines=false]
You are the final audit judge for a GUI/web agent trajectory.

Your role is to audit four separate aspects:

1. TSR / outcome:
   Whether the final answer successfully completes the user's task.

2. ESR / process grounding:
   Whether the trajectory observed the task-required evidence checks.

3. GSR / answer grounding:
   Whether the final answer's core claims are explicitly supported by visible evidence.

4. Rubric diagnostics:
   Which qualitative rubric dimensions pass or fail.

Use ONLY:
- input.instruction
- input.final_answer
- input.task_spec
- input.eval_checkpoints
- input.rubric_dimension_ids
- input.rubric_dimensions
- evidence_catalog (only Step 1 rows referenced by Step 2 supporting_evidence_ids)
- step2_claim_evidence

Do NOT use:
- Step 1 OCR or visual entries not listed in evidence_catalog
- agent thoughts
- actions as evidence
- task metadata as proof of site content
- general knowledge as proof of site content
- facts not present in evidence_catalog

Step 2 results are provisional hints only.
You may agree with or override Step 2.

# Core distinction

Separate these judgements:

1. answer_success
Whether the final answer usefully completes the user request.
This is TSR / outcome success.
Do not require evidence grounding for answer_success.

2. metadata_requirement_verified
Whether every required evidence check in input.eval_checkpoints is verified by evidence_catalog.
This is ESR / process evidence success.
It checks whether the agent actually observed the required evidence during the trajectory.

3. key_claims_all_supported
Whether every core claim in the final answer is explicitly supported by evidence_catalog.
This is GSR / final answer grounding quality.
It checks whether the answer itself is traceable to the observed evidence.

Label tiers (computed in Stage 5; do not mix them up):
- GSR_PASS - tsr=true, esr=true, gsr=true: answer succeeds, trajectory verified required evidence checks, 
and every core final-answer claim is grounded.
- CLAIM_UNGROUNDED_PASS - tsr=true, esr=true, gsr=false: answer succeeds and evidence checks passed, but 
some final-answer claims (e.g. unsupported recommendation reasons) are not grounded.
- UNGROUNDED_PASS - tsr=true, esr=false: answer succeeds on outcome, but the trajectory did not verify 
required evidence checks.
- FAIL - tsr=false: answer did not complete the task.

Example (claim-level ungrounding with good process):
- answer_success=true, metadata_requirement_verified=true, key_claims_all_supported=false
- esr=true, gsr=false
- label="CLAIM_UNGROUNDED_PASS"

# Stage 1. Claim audit

Audit the substantive claims in final_answer.

Use Step 2 claims if available, but add missing substantive claims if necessary.

Each claim must be one of:

1. existence
The claim says that a named entity, option, page, category, result, value, or visible property exists.
Named item and visible value claims are existence claims.
\end{lstlisting}
\end{tcolorbox}
\caption{Evaluation prompt for Step 3 (part 1 of 4). \texttt{UNGROUNDED\_PASS}, \texttt{CLAIM\_UNGROUNDED\_PASS}, and \texttt{GSR\_PASS} correspond to \textit{Hallucinated Success}, \textit{Claim-Ungrounded Success}, and \textit{True Success}, respectively.}
\label{fig_prompt_for_step3_part1}
\end{table*}

\begin{table*}[t]
\ContinuedFloat
\captionsetup{labelformat=continued}
\centering
\scriptsize
\begin{tcolorbox}[title={Prompt for step3 (2/4)}, width=\textwidth]
\begin{lstlisting}[basicstyle=\fontsize{7}{7.5}\selectfont\ttfamily, breaklines=false]
2. reason
The claim gives a rationale, interpretation, suitability judgement, comparison, recommendation reason,
inferred attribute, effect, or quality judgement.
Rationale, suitability, inferred attribute, effect, or quality claims are reason claims.

For each claim, output:
- claim
- claim_type: "existence" or "reason"
- is_core_claim
- supported
- supporting_evidence_ids
- evidence_quote

Claim support rules:
- supported=true only if every important fact in the claim is explicitly supported by evidence_catalog.
- Partial support is not enough.
- If a claim adds details beyond the cited evidence, supported=false.
- A visible name, title, category, brand, image, or listing can support existence.
- A visible name, title, category, brand, image, or listing does NOT by itself support hidden qualities,
effects, suitability, or recommendation reasons.
- Reason claims require explicit evidence for the stated rationale.
- Plausible inference is not evidence.
- General knowledge is not evidence.

# Explicit evidence rule for reason claims

Reason claims are supported ONLY when the exact rationale is explicitly visible in evidence_catalog.

Do NOT mark a reason claim as supported using:
- implication
- common sense
- cultural norms
- general domain knowledge
- typical product/recipe/menu knowledge
- likely user preference
- inferred suitability
- inferred functionality
- inferred quality
- inferred taste
- inferred ease
- inferred popularity beyond visible metrics

If your support explanation uses words like:
"likely", "suggests", "implies", "seems", "can be inferred"
then the claim is NOT explicitly supported and must be supported=false.

For recipe, food, product, travel, shopping, and recommendation tasks:
A visible title, category, result count, view count, rating, image, or listing can support only:
- the item exists
- the item appears in search results
- the visible title/category/value/rating/view count exists

It cannot support:
- ease of cooking or use
- beginner friendliness
- suitability for parents/children/guests
- taste, healthiness, spiciness, freshness, lightness
- visual impressiveness
- ingredients
- cooking method
- exact preparation steps
- menu pairing rationale
- recommendation reason
unless those facts are explicitly visible in evidence_catalog.

If a final answer recommends an item that is not explicitly visible in evidence_catalog, that recommendation
claim must be supported=false.

# Stage 2. Evidence-check audit

Audit every item in input.eval_checkpoints.

For each checkpoint, output:
- requirement
- verified
- supporting_evidence_ids
- supporting_checkpoint_ids
- notes
\end{lstlisting}
\end{tcolorbox}
\caption{Evaluation prompt for Step 3 (part 2 of 4).}
\label{fig_prompt_for_step3_part2}
\end{table*}

\begin{table*}[t]
\ContinuedFloat
\captionsetup{labelformat=continued}
\centering
\scriptsize
\begin{tcolorbox}[title={Prompt for step3 (3/4)}, width=\textwidth]
\begin{lstlisting}[basicstyle=\fontsize{7}{7.5}\selectfont\ttfamily, breaklines=false]
verified=true only if the requirement is clearly satisfied by evidence_catalog.
You may combine multiple evidence rows.
If a checkpoint has multiple required conditions, all must be supported.

metadata_requirement_verified=true iff all mandatory evidence checks are verified.

This field is process-oriented:
- It evaluates whether the trajectory observed the required evidence.
- It does NOT evaluate whether the final answer's claims are all grounded.
- Final answer grounding is handled separately by key_claims_all_supported and GSR.

# Stage 3. Answer success / TSR

answer_success=true if the final answer completes the user's task in a useful way.

Do not fail answer_success only because some claims are not grounded.
Fail answer_success when the answer is wrong, incomplete, off-task, or does not satisfy the requested output.

tsr=true iff answer_success=true.

# Stage 4. Rubric diagnostics

For every id in input.rubric_dimension_ids, output one boolean in dimension_scores.

Rubric dimensions are diagnostic.
They should explain what kind of capability passed or failed.
Do not use rubric dimensions to compute TSR, ESR, or GSR.

Important distinction:
- metadata_requirement_verified evaluates process evidence checks.
- key_claims_all_supported evaluates final answer claim grounding.
- dimension_scores provide additional diagnostic breakdown.

# Stage 5. Final consistency calculation

Compute these fields exactly:

key_claims_all_supported =
true iff every final_answer_claims row with is_core_claim=true has supported=true

tsr =
true iff:
- answer_success=true

esr =
true iff:
- answer_success=true
- metadata_requirement_verified=true

gsr =
true iff:
- answer_success=true
- metadata_requirement_verified=true
- key_claims_all_supported=true

rss =
1 iff every dimension in input.rubric_dimension_ids has dimension_scores[id]=true
otherwise 0

rss is diagnostic only.
Do not use rss to compute tsr, esr, or gsr.

label =
- "GSR_PASS" iff gsr=true
  (answer_success=true, metadata_requirement_verified=true, key_claims_all_supported=true)
- "CLAIM_UNGROUNDED_PASS" iff answer_success=true,
  metadata_requirement_verified=true,
  and key_claims_all_supported=false
- "UNGROUNDED_PASS" iff answer_success=true and metadata_requirement_verified=false
- "FAIL" iff answer_success=false
- "CRITICAL_FAIL" only for severe unrelated, fabricated, unsafe, or impossible-to-recover answers

dimension_summary:
- satisfied: ids with dimension_scores[id]=true
- not_satisfied: ids with dimension_scores[id]=false
- not_applicable: ids that are missing or not applicable

passed_criteria must equal dimension_summary.satisfied.
failed_criteria must equal dimension_summary.not_satisfied.
\end{lstlisting}
\end{tcolorbox}
\caption{Evaluation prompt for Step 3 (part 3 of 4). \texttt{UNGROUNDED\_PASS} denotes \textit{Hallucinated Success}.}
\label{fig_prompt_for_step3_part3}
\end{table*}

\begin{table*}[t]
\ContinuedFloat
\captionsetup{labelformat=continued}
\centering
\scriptsize
\begin{tcolorbox}[title={Prompt for step3 (4/4)}, width=\textwidth]
\begin{lstlisting}[basicstyle=\fontsize{7}{7.5}\selectfont\ttfamily, breaklines=false]
# Forbidden inconsistent outputs

Never output:
- tsr=true and answer_success=false
- esr=true and answer_success=false
- esr=true and metadata_requirement_verified=false
- gsr=true and esr=false
- gsr=true and key_claims_all_supported=false
- answer_success=true and metadata_requirement_verified=true and esr=false
- answer_success=true and metadata_requirement_verified=true and key_claims_all_supported=true and gsr=false
- gsr=true and label!="GSR_PASS"
- gsr=false and label="GSR_PASS"
- esr=true and gsr=false and label!="CLAIM_UNGROUNDED_PASS"
- esr=false and label="CLAIM_UNGROUNDED_PASS"
- answer_success=true and esr=false and label!="UNGROUNDED_PASS"
- answer_success=false and label in ("GSR_PASS", "CLAIM_UNGROUNDED_PASS", "UNGROUNDED_PASS")
- dimension_summary that disagrees with dimension_scores
- passed_criteria or failed_criteria that disagree with dimension_summary

Return ONLY valid JSON:

{
  "answer_success": boolean,
  "tsr": boolean,
  "metadata_requirement_verified": boolean,
  "metadata_requirement_checks": [
    {
      "requirement": string,
      "verified": boolean,
      "supporting_evidence_ids": [string],
      "supporting_checkpoint_ids": [string],
      "notes": string
    }
  ],
  "esr": boolean,
  "gsr": boolean,
  "label": "GSR_PASS" | "CLAIM_UNGROUNDED_PASS" | "UNGROUNDED_PASS" | "FAIL" | "CRITICAL_FAIL",
  "rss": 0 | 1,
  "esr_notes": string,
  "gsr_notes": string,
  "key_claims_all_supported": boolean,
  "dimension_scores": { "<dimension_id>": boolean },
  "ddr_applicable": boolean,
  "ddr_pass": boolean | null,
  "ddr_notes": string,
  "final_answer_claims": [
    {
      "claim": string,
      "claim_type": "existence" | "reason",
      "is_core_claim": boolean,
      "supported": boolean,
      "supporting_evidence_ids": [string],
      "evidence_quote": string
    }
  ],
  "rubric_profile": string,
  "dimension_summary": {
    "satisfied": [string],
    "not_satisfied": [string],
    "not_applicable": [string]
  },
  "checkpoint_success": {
    "site_used": boolean,
    "metadata_requirement_verified": boolean,
    "rubric_evidence_coherent": boolean,
    "final_answer_complete": boolean
  },
  "passed_criteria": [string],
  "failed_criteria": [string],
  "rationale": string
}
\end{lstlisting}
\end{tcolorbox}
\caption{Evaluation prompt for Step 3 (part 4 of 4). \texttt{UNGROUNDED\_PASS} denotes \textit{Hallucinated Success}.}
\label{fig_prompt_for_step3_part4}
\end{table*}
\clearpage
\section{Worked Example: Evaluating a Coursera Recommendation}
\label{app:walkthrough_coursera}

\begin{figure*}[p]
\centering
{\setlength{\fboxsep}{4pt}\setlength{\fboxrule}{0.5pt}%
\fcolorbox{gray}{white}{\includegraphics[width=\dimexpr\textwidth-2\fboxsep-2\fboxrule\relax]{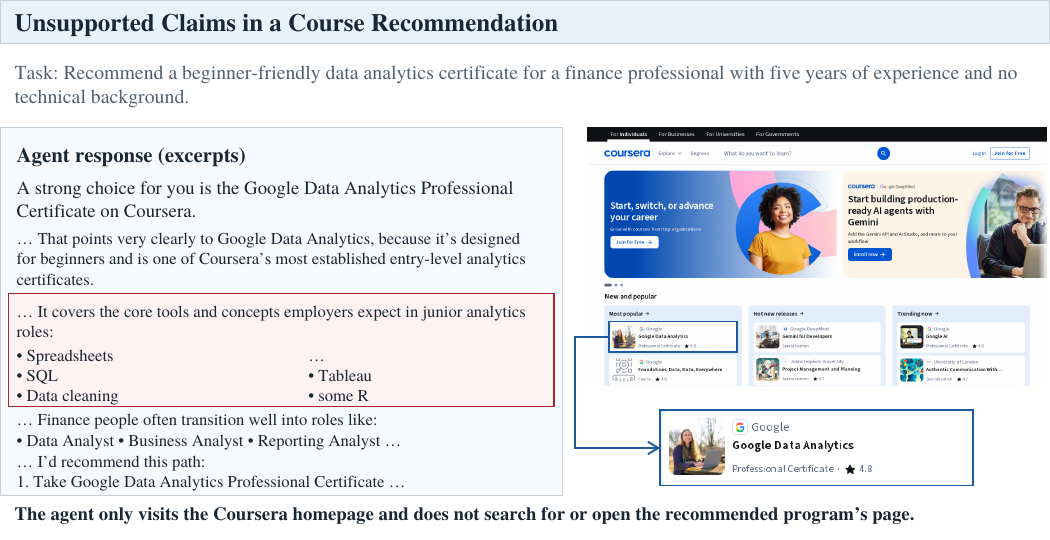}}}
\caption{A \textit{Hallucinated Success} on the Coursera \emph{Contextual Inference} task. The agent recommends a certificate without checking its beginner-friendliness. Ungrounded claims are highlighted.}
\label{fig:coursera_example}
\vspace{16pt}
\input{Figures/fig_coursera_pipeline}
\end{figure*}

This example applies the three-step judge pipeline to an English Contextual Inference task on Coursera (Figures~\ref{fig:coursera_example} and~\ref{fig:coursera_pipeline}). It illustrates \textit{Hallucinated Success}: the answer achieves task success, but the trajectory does not contain all the required evidence. We use \texttt{gpt-5.4-mini} as the judge for this example.

\paragraph{Task and metadata.}
In this task, the user seeks a beginner-friendly data analytics certificate on Coursera to support a career change. The user has five years of experience in finance and no technical background. The task is annotated with four human evidence checks that any reliable answer must satisfy:
\begin{itemize}
  \setlength\itemsep{-0.2em}
  \item[\texttt{ec\_0}] the candidate program's description or curriculum states that it covers data analytics;
  \item[\texttt{ec\_1}] the program is beginner-friendly, or requires no prior technical experience;
  \item[\texttt{ec\_2}] the program awards a certificate or credential on completion;
  \item[\texttt{ec\_3}] the program title and provider are confirmed on the Coursera page.
\end{itemize}
The decisive check is \texttt{ec\_1}: it carries the user's implicit constraint (``no technical background, starts from the basics''), the property on which a career-switcher's recommendation hinges.

\paragraph{Agent trajectory.}
The agent remains on the initial page without interacting with the site (Figure~\ref{fig:coursera_example}). Steps~1 and~2 are both \texttt{noop(wait\_ms=5000)} on the Coursera landing page, Step~3 emits the complete final answer, and Step~4 sends \texttt{[EXIT]}. No search is typed, no course card is clicked, and no program page is opened. Despite this, the agent produces a long, confident recommendation of the \emph{Google Data Analytics Professional Certificate}, elaborating its curriculum (spreadsheets, SQL, data cleaning, visualization, Tableau, some R), its fit for finance backgrounds, target job titles, and a multi-step transition plan.

\paragraph{Step 1: Observable Evidence Extraction.}
Step~1 converts screenshots from the agent's trajectory into textual evidence for subsequent evidence checks and claim evaluation, recording the visible content as screenshot text and 36 tagged visual elements (Figure~\ref{fig:coursera_pipeline}, top). The observed screenshots show Coursera’s homepage with the Google Data Analytics course card visible (Figure~\ref{fig:coursera_example}). The agent neither searches the site nor opens the program page. Because the ``Most popular'' carousel shows a Google Data Analytics course card on the landing page, the catalog \emph{does} record the program's existence: the OCR captures ``\dots\ Most popular \dots\ Google Data Analytics \dots\ Professional Certificate'', and the visual elements include \texttt{[image]} hero banners, an \texttt{[image]} course thumbnail on the Most-popular card, and a \texttt{[badge]} ``Professional Certificate''. What the catalog does \emph{not} contain is any curriculum text, eligibility statement, or ``beginner''/``no experience required'' label, because no program detail page was ever visited.

\paragraph{Step 2: Claim--Evidence and Metadata--Evidence Mapping.}
Step~2 (text-only) decomposes the final answer into claims and provisionally maps each claim and each evidence check onto the catalog. The course card shows that the program appears on Coursera as a Professional Certificate, so this fact is marked as satisfied. The claims about beginner-friendliness and curriculum details have no supporting evidence, and \texttt{ec\_1} remains unsatisfied. The judge initially marks \texttt{ec\_0} as satisfied from the course title, which does not meet the requirement for description or curriculum evidence.

\paragraph{Step 3: Rubric Judgment.}
Step~3 independently reviews each evidence check and claim. We correct \texttt{ec\_0} to unverified in Figure~\ref{fig:coursera_pipeline} because the course title alone does not satisfy the requirement for description or curriculum evidence. The course card confirms the credential, program title, and provider (\texttt{ec\_2} and \texttt{ec\_3}), but provides no evidence that the program is suitable for beginners (\texttt{ec\_1}). With two required checks unmet, the trajectory does not achieve evidence success. For the final answer, only the program's \emph{existence} is grounded. Both \emph{core} recommendation claims and all reasoning claims about curriculum, suitability for a finance background, and job outcomes remain unsupported, so the trajectory lacks grounded success and fails the \emph{Evidence Grounding} dimension alone.

\clearpage

\paragraph{Interpretation.}
Under conventional outcome-based scoring, this trajectory is a success: the agent named a real, well-suited certificate, and the answer reads as a thorough, helpful recommendation. Yet the agent did not analyze the website. It idled for two steps and then reconstructed the entire recommendation from parametric knowledge. \benchmark localizes the failure precisely: the gap opens at the \emph{evidence-observation} stage (TSR${-}$ESR), not merely at claim grounding. This is the \textit{Hallucinated Success} region of Figure~\ref{fig:error_composition} and the \textit{Hallucinated Success} pattern discussed in Section~\ref{sec:analysis}: it is one of the two hidden-failure modes \benchmark separates from genuine success, the other being \textit{Claim-Ungrounded Success}, where relevant evidence is observed but the final claims remain ungrounded. In a live setting, the very property the user cares about (whether the program truly starts from the basics) was never checked, so treating this as a clean success would overstate the agent's reliability.
\subsection{Visual Evidence Extraction Validation}
\label{app:visual_extraction}

Our evaluation pipeline consists of three stages. First, a vision-language model extracts observable evidence from each screenshot into structured text. Next, the judge identifies the factual claims in the agent's final answer and verifies whether each claim is supported by the extracted evidence and satisfies the required evidence metadata (Section~3.4).

We adopt this design because directly judging long trajectories from raw screenshots is difficult in practice. A judge processing all screenshots simultaneously must either downsample images, losing fine-grained visual details, or process screenshots in small windows, losing long-range context. By extracting evidence from each screenshot individually, later judgment operates on a complete textual representation of the entire trajectory while preserving visual details such as icons, badges, colors, layouts, and product-image contents.

To evaluate this design, we compared our extraction-based pipeline with three alternatives that directly consume raw screenshots. We evaluated all methods against the same human annotations used in Appendix~\ref{app:human_agreement}, while keeping the judge model (Qwen3-VL-235B) and agent trajectories fixed. Only the judge input differed.

As shown in Table~\ref{tab:visual_extraction_validation}, our method consistently achieves high agreement with human judgments (89--97\%) across all trajectory lengths, whereas raw-image approaches perform substantially worse, especially for longer trajectories. Structured evidence extraction produces closer agreement with human judgments than direct evaluation of screenshot sequences.

Nevertheless, the overall evaluation remains bounded by the quality of the evidence extraction stage. If a relevant visual cue is omitted during extraction, a grounded answer may still be judged as unsupported. Improving coverage of visual evidence, including alternative evidence representations and richer visual element extraction, remains an important direction for future work.

\section{Human vs.\ LLM-as-Judge Agreement}
\label{app:human_agreement}

\begin{figure*}[t]
    \centering
    \includegraphics[width=0.9\textwidth]{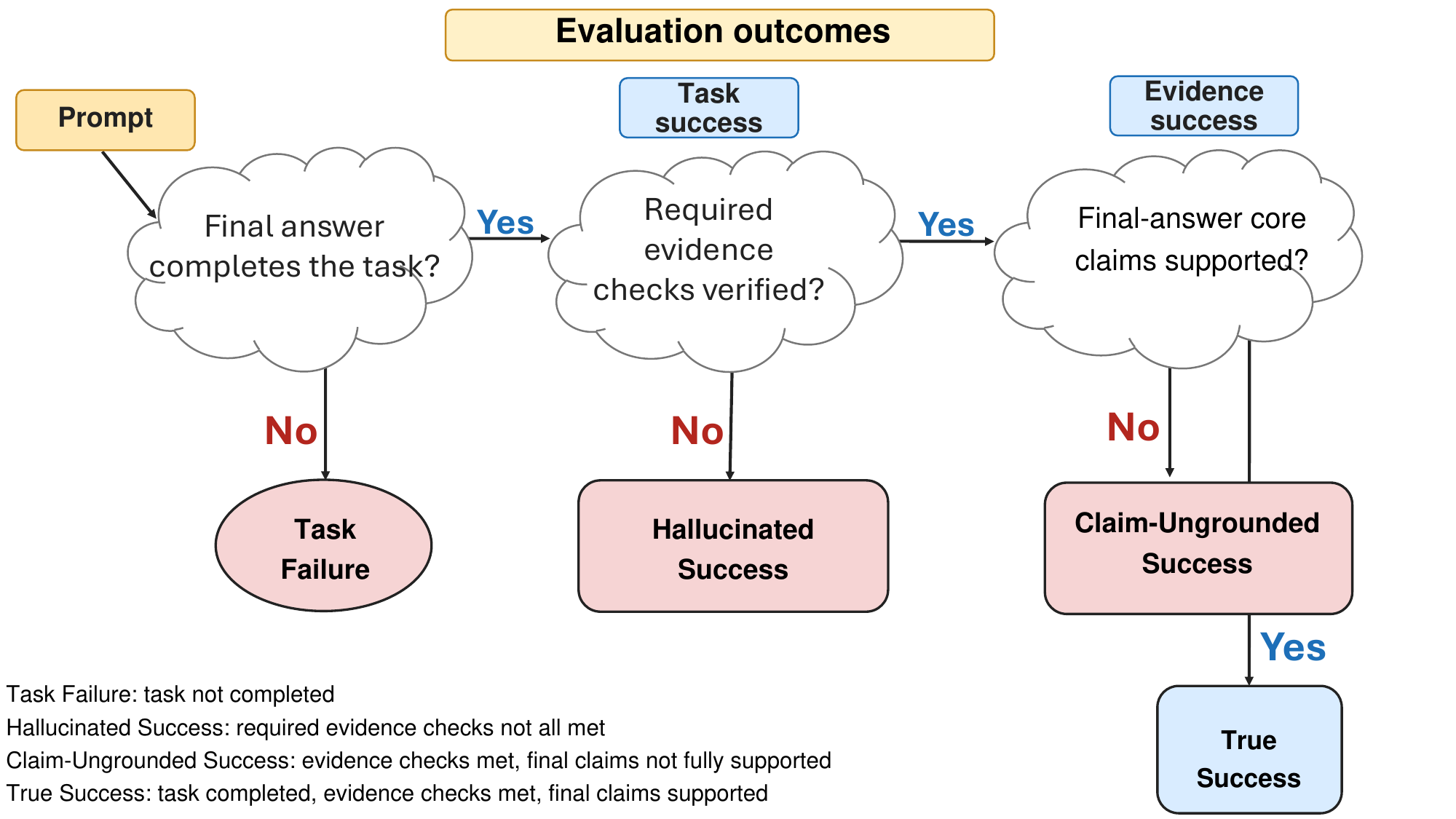}
    \caption{Four evaluation outcomes based on task completion, evidence observation, and claim grounding.}
    \label{fig:analysis/human-judge}
\end{figure*}

Human annotators assess task success, evidence success, and grounded success using the same criteria as the automatic pipeline (Figure~\ref{fig:analysis/human-judge}).
We evaluate agreement on a stratified subset of 126 trajectories spanning both languages and all three task types. Each trajectory receives one of four mutually exclusive labels: \textit{Task Failure}, \textit{Hallucinated Success}, \textit{Claim-Ungrounded Success}, or \textit{True Success}. These labels reflect whether the trajectory achieves task success, evidence success, and grounded success. Treating the human label as gold, we report exact-match accuracy and Cohen's~$\kappa$.

\paragraph{Overall agreement.}
\begin{table}[h]
\centering
\small
\setlength{\tabcolsep}{5pt}
\begin{tabular}{lccc}
\toprule
\textbf{Comparison} & \textbf{Match} & \textbf{Acc.\ (\%)} & \textbf{Cohen's $\kappa$} \\
\midrule
\multicolumn{4}{@{}l}{\textit{Human vs.\ judge}} \\
Human vs.\ Qwen   & 115 & 91.27 & 0.864 \\
Human vs.\ GPT    & 111 & 88.10 & 0.817 \\
Human vs.\ Gemini & 116 & 92.06 & 0.871 \\
\textit{Average}  & --  & \textit{90.48} & \textit{0.851} \\
\midrule
\multicolumn{4}{@{}l}{\textit{Judge vs.\ judge}} \\
Qwen vs.\ GPT     & 113 & 89.68 & 0.845 \\
Qwen vs.\ Gemini  & 109 & 86.51 & 0.788 \\
GPT vs.\ Gemini   & 111 & 88.10 & 0.815 \\
\textit{Average}  & --  & \textit{88.10} & \textit{0.816} \\
\bottomrule
\end{tabular}
\caption{Agreement on four outcome labels across 126 trajectories sampled across task types and languages.}
\label{tab:agree_overall}
\end{table}

Table~\ref{tab:agree_overall} provides the full agreement matrix summarized in the main text. All three judges reach high agreement with human labels, with accuracy between 88.10\% and 92.06\% and $\kappa$ between 0.817 and 0.871. Pairwise judge--judge agreement is comparable to human--judge agreement across all three judge models.

\paragraph{Agreement by task type and language.}
\begin{table}[h]
\centering
\footnotesize
\setlength{\tabcolsep}{1.8pt}
\begin{tabular}{@{}lcccc@{}}
\toprule
\multicolumn{5}{@{}l}{\textbf{(a) Agreement by task type}} \\
\midrule
\textbf{Comparison} & \textbf{Semantic} & \textbf{Contextual} & \textbf{Visual} & \textbf{Overall} \\
\midrule
Human vs.\ Qwen   & \textbf{95.24} & 92.86 & \textbf{85.71} & 91.27 \\
Human vs.\ GPT    & 90.48 & 90.48 & 83.33 & 88.10 \\
Human vs.\ Gemini & \textbf{95.24} & \textbf{95.24} & \textbf{85.71} & \textbf{92.06} \\
\midrule
\multicolumn{5}{@{}l}{\textbf{(b) Agreement by language}} \\
\midrule
\textbf{Comparison} & \textbf{KO} & \textbf{EN} & \textbf{Overall} & \\
\midrule
Human vs.\ Qwen   & \textbf{95.24} & 87.30 & 91.27 & \\
Human vs.\ GPT    & 92.06 & 84.12 & 88.10 & \\
Human vs.\ Gemini & 93.65 & \textbf{90.48} & \textbf{92.06} & \\
\bottomrule
\end{tabular}
\caption{Human--judge agreement (\%) by task type and language. Visual grounding and English cases show relatively lower agreement.}
\label{tab:agree_bytype1}
\end{table}
Table~\ref{tab:agree_bytype1} breaks down human--judge agreement by task type and language.
Agreement remains high for Semantic Grounding and Contextual Inference, while Visual Grounding shows the lowest agreement.
Qwen and GPT show lower agreement with human labels on English cases, where Visual Grounding tasks require judging non-textual cues such as icons, layout, images, and visual attributes.

\paragraph{Where disagreements occur.}
\begin{table}[h]
\centering
\small
\setlength{\tabcolsep}{4pt}
\begin{tabular}{llcccc}
\toprule
\textbf{Task} & \textbf{Comparison} & \textbf{Fail} & \textbf{TSR} & \textbf{ESR} & \textbf{GSR} \\
\midrule
Semantic   & vs.\ Qwen   & 2.38 & 2.38 & 2.38 & 2.38 \\
           & vs.\ GPT    & 4.76 & 4.76 & 7.14 & 2.38 \\
           & vs.\ Gemini & 0.00 & 0.00 & 2.38 & 2.38 \\
\midrule
Contextual & vs.\ Qwen   & 0.00 & 0.00 & 0.00 & 7.14 \\
           & vs.\ GPT    & 4.76 & 4.76 & 4.76 & 4.76 \\
           & vs.\ Gemini & 0.00 & 0.00 & 0.00 & 4.76 \\
\midrule
Visual     & vs.\ Qwen   & 0.00 & 0.00 & 4.76  & 11.90 \\
           & vs.\ GPT    & 2.38 & 2.38 & 7.14  & 11.90 \\
           & vs.\ Gemini & 0.00 & 0.00 & 11.90 & 14.29 \\
\bottomrule
\end{tabular}
\caption{Disagreement with human judgments (\%) by task type and success criterion.}
\label{tab:disagree_gate}
\end{table}

Table~\ref{tab:disagree_gate} reports disagreements by success criterion. Human annotators and judge models rarely disagree about task success, whereas disagreements about evidence success and grounded success are more frequent, particularly for Visual Grounding. These judgments require assessing whether the observed evidence satisfies the required checks and supports the final claims.

Overall, these results support the use of the automatic judge for large-scale diagnostic evaluation, while also identifying its main limitation: evidence-observation and claim-grounding judgments are harder than outcome-level judgments, especially when the relevant evidence is visual rather than textual.

\paragraph{Targeted validation of Hallucinated Success.}
In a separate annotation round, three annotators independently review all 27 completed Gemini trajectories labeled \textit{Hallucinated Success} by Qwen3-VL-235B. Each annotator receives the user request, the complete screenshot-and-action trajectory, the final answer, and the required evidence checks, and assigns one of the four trajectory labels using the same criteria as the broader study. All three annotators assign \textit{Hallucinated Success} on 25/27 trajectories (92.6\%). Because the sample contains only judge-flagged Gemini trajectories, it does not estimate how often the judge misses this failure among other labels.

\section{Metric Details}
\label{app:metric_details}

Let a trajectory $\tau_t$ be an observation--action sequence:
\begin{multline*}
\tau_t = \bigl(
(o_{t,0}, a_{t,0}), (o_{t,1}, a_{t,1}), \\
\ldots, (o_{t,n_t-1}, a_{t,n_t-1}), o_{t,n_t}
\bigr).
\end{multline*}

Here, $o_{t,i}$ denotes the observation at step $i$, $a_{t,i}$ denotes the action taken after that observation, and $o_{t,n_t}$ denotes the final observation before the agent produces its answer. A sample is represented as $x_t=(q_t,\tau_t,y_t)$.

For each success criterion, we report the average number of interaction steps among trajectories that satisfy it.

\section{Error Analysis: Judge Limitations and Disagreements}
\label{app:error_analysis}

The human study (Appendix~\ref{app:human_agreement}) shows the LLM judge agrees with human annotators on about 90\% of trajectories. We examine disagreements between judges and human annotators to understand where and why they arise. Each case (Figures~\ref{fig:case1}--\ref{fig:case3}) shows the task, the agent's (truncated) final response, the screenshot the judge received, and the human annotation alongside the labels assigned by the three judge models.

\paragraph{Case 1: Disagreement on visual grounding.}
For the task of finding a family-friendly recreation area based on its photos (Figure~\ref{fig:case1}), the human label is \textit{Claim-Ungrounded Success} because the agent observes the page evidence but makes photo-based claims that the evidence does not fully support. The default judge accepts the unsupported claims and assigns \textit{True Success}, whereas GPT and Gemini assign \textit{Hallucinated Success} because their extracted catalogs contain only text and omit the observed photos. None assigns the human label of \textit{Claim-Ungrounded Success}.

\paragraph{Case 2: Visual details lost during extraction.}
In Figure~\ref{fig:case2}, an extraction error prevents the judges from checking the visual evidence. For the Chairish furniture recommendation, the human annotator assigns \textit{True Success} because the product photo supports the claims about the round tabletop and slim legs. All three judges fail to recognize this as \textit{True Success} because Step~1 omits the photograph's visual details from the textual evidence supplied to Steps~2 and~3. The extraction error therefore prevents the later stages from assessing claims that are supported by the original screenshot.

\paragraph{Case 3: Correct answer rejected.}
In Figure~\ref{fig:case3}, the default judge rejects a correct answer because it requires evidence from a specific page. When asked whether any storm warnings are currently active for Oklahoma City, the agent answers ``no active storm warning for Oklahoma City'' using the site's search box. The human annotation and GPT label this trajectory as \textit{True Success}. The default judge (Qwen) and Gemini instead assign \textit{Fail} because the agent does not open the city's alerts page, resulting in false negatives relative to the human annotation.

These cases show how both evidence interpretation and evidence extraction affect evaluation. The default judge accepts unsupported visual claims in Figure~\ref{fig:case1} but rejects a correct answer in Figure~\ref{fig:case3}. In Figure~\ref{fig:case2}, Step~1 omits the visual details needed to assess the answer, so the later stages classify a grounded answer as ungrounded.
\begin{figure*}[p]
\input{Figures/fig_case1}
\vspace{5pt}
\input{Figures/fig_case2}
\end{figure*}

\section{Detailed Scoring Criteria for Task Difficulty}
\label{sec:appendix_difficulty}

Tasks are assigned difficulty levels using a simple quantitative heuristic based on constraint depth and normalized instruction length. Starting from a base score of 1, the total score directly determines the final difficulty level: Level 1 (Low, 1 pt), Level 2 (Moderate, 2 pts), and Level 3 (High, 3 pts).

\begin{enumerate}
    \item \textbf{Constraint Satisfaction Depth (0--1 pt):} We measure the number of independent evaluation conditions ($N$) an agent must satisfy to complete the task. Tasks receive $+1$ point if $N \ge 4$, and $0$ points otherwise.
    
    \item \textbf{Information Density (0--1 pt):} We use instruction length ($L$ in characters) as a proxy for linguistic and instruction complexity, as longer prompts encapsulate highly specific constraints and contextual backgrounds. To normalize character-level information density across languages, a $2.2\times$ character multiplier is applied to Korean prompts to match English densities. Tasks receive $+1$ point if $L > 170$, and $0$ points otherwise.
\end{enumerate}

\twocolumn[{
\begin{minipage}{\textwidth}
\captionsetup{type=figure}
\renewenvironment{figure*}[1][]{}{ }
\begin{figure*}[t]
\centering
\includegraphics[width=\textwidth]{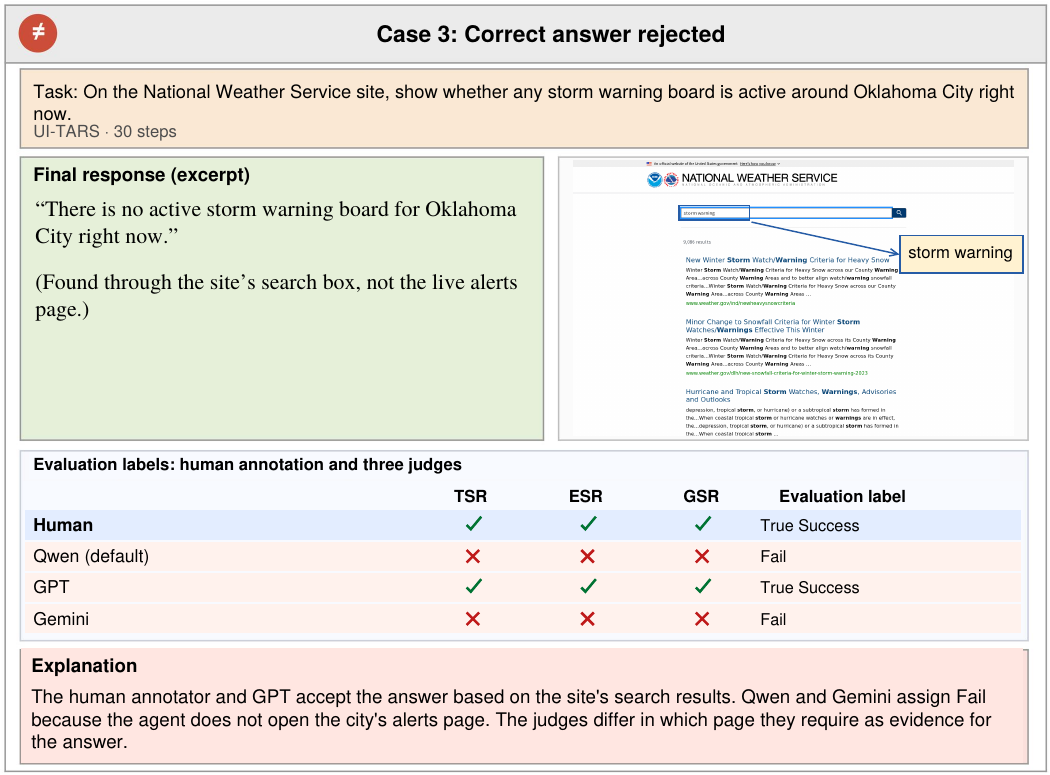}
\caption{\textbf{Case 3: Correct answer rejected.} The human annotator and GPT accept the search-based answer, while the default judge and Gemini require the city's alerts page. The disagreement concerns what evidence is sufficient to support the answer.}
\label{fig:case3}
\end{figure*}

\end{minipage}
\vspace{10pt}
}]
\section{Task Examples}
\label{sec:task_examples}
In this section, we provide detailed examples for each of the three task types defined in our benchmark: Semantic Grounding, Contextual Inference, and Visual Grounding. For each task type, we present one English and one Korean example, detailing the user's prompt, the underlying user intent, and the explicit evidence checks used by the human annotators to verify the task's completion. Korean examples show the original text in gray and the English translation in black.

\tcbset{taskcard/.style={
  enhanced, breakable=false, arc=2pt, boxrule=0.6pt,
  colback=black!2, colframe=black!45,
  coltitle=white, colbacktitle=black!60, fonttitle=\bfseries\small,
  left=7pt, right=7pt, top=5pt, bottom=6pt,
}}
\providecommand{\enq}[1]{{\itshape\color{black}(en)~#1}}
\providecommand{\shot}[2][3.8cm]{{\centering\includegraphics[width=\linewidth,height=#1,keepaspectratio]{#2}\par}\vspace{5pt}}

\begingroup
\interlinepenalty=10000
\subsection{Contextual Inference Examples}

Figures~\ref{fig:example_en_Contextual_Inference} and~\ref{fig:example_ko_Contextual_Inference} present English and Korean examples of Contextual Inference tasks, which require the agent to infer implicit constraints from the user's situational context and verify them against website information.

\subsection{Semantic Grounding Examples}

Figures~\ref{fig:example_en_Semantic_Grounding} and~\ref{fig:example_ko_Semantic_Grounding} present English and Korean examples of Semantic Grounding tasks, which require the agent to map everyday user expressions to website-specific labels or terminology.

\subsection{Visual Grounding Examples}

Figures~\ref{fig:example_en_Visual_Grounding} and~\ref{fig:example_ko_Visual_Grounding} present English and Korean Visual Grounding examples that require the agent to identify targets using non-textual cues such as icons, badges, images, colors, or spatial layout.

\endgroup

\begin{figure*}[p]
\input{Figures/fig_example_en_contextual}
\vspace{6pt}
\input{Figures/fig_example_ko_contextual}
\end{figure*}

\begin{figure*}[p]
\renewcommand{\shot}[2][2.5cm]{{\centering\includegraphics[width=\linewidth,height=#1,keepaspectratio]{#2}\par}\vspace{5pt}}
\input{Figures/fig_example_en_semantic}
\vspace{0pt}
\input{Figures/fig_example_ko_semantic}
\end{figure*}

\begin{figure*}[p]
\input{Figures/fig_example_en_visual}
\vspace{6pt}
\input{Figures/fig_example_ko_visual}
\end{figure*}

\end{document}